\documentclass[review]{elsarticle}
\usepackage{lineno,hyperref}

\usepackage[english]{babel}
\usepackage{amsmath, amsxtra, amsfonts, amssymb, amstext}
\usepackage{amsthm}
\usepackage{booktabs}
\usepackage{nicefrac}
\usepackage{xspace}
\usepackage{url}
\usepackage{graphics,color}
\definecolor{linkblue}{rgb}{0.1,0.1,0.8}
\hypersetup{colorlinks=true,linkcolor=linkblue,filecolor=linkblue,urlcolor=linkblue,citecolor=linkblue}
\usepackage[algo2e,ruled,vlined,linesnumbered]{algorithm2e}
\usepackage{multirow}
\newcommand{\assign}{\leftarrow}
\usepackage{pdfpages}
\usepackage{bm}
\usepackage{enumerate}
\usepackage{mathtools}

\usepackage{array}
\usepackage{booktabs}
\usepackage{bigdelim}

\newtheorem{theorem}{Theorem}
\newtheorem{lemma}[theorem]{Lemma}

\newcommand{\N}{\mathbb{N}}
\newcommand{\R}{\mathbb{R}}

\renewcommand{\epsilon}{\varepsilon}
\newcommand{\eps}{\varepsilon}

\DeclareMathOperator{\E}{\mathbb{E}}

\DeclareMathOperator{\rk}{rk}

\DeclarePairedDelimiter{\ceil}{\lceil}{\rceil}
\DeclarePairedDelimiter{\floor}{\lfloor}{\rfloor}

\newcommand{\ooea}{$(1 + 1)$-EA\xspace}
\newcommand{\olea}{$(1 + \lambda)$-EA\xspace}
\newcommand{\moea}{$(\mu + 1)$-EA\xspace}
\newcommand{\moga}{$(\mu + 1)$-GA\xspace}
\newcommand{\ollga}{$(1 + (\lambda,\lambda))$-GA\xspace}

\newcommand{\onemax}{\textsc{OneMax}\xspace}
\newcommand{\OM}{\textsc{Om}\xspace}

\newcommand{\Binval}{\textsc{BinVal}\xspace}

\newcommand{\hottopic}{\textsc{HotTopic}\xspace}
\newcommand{\HT}{\textsc{HT}\xspace}

\newcommand{\Ae}{A_\ell}
\newcommand{\Rel}{R_\ell}

\newcommand{\best}[1]{Z_{#1}}
\newcommand{\tbirth}[1]{t_{#1}}
\newcommand{\tdie}[1]{T_{#1}}
\newcommand{\ac}{\alpha c}

\newcommand{\Egood}{\mathcal E_{\textup{good}}}

\newcommand{\etmac}{e^{-\ac}}
\newcommand{\etac}{e^{\ac}}

\newcommand{\mlm}{\mu \log\mu}
\newcommand{\lm}{\log\mu}
\newcommand{\Xgi}{X_{\ge i}}

\newcommand{\event}[1]{\mathcal E_{#1}}

\journal{Theoretical Computer Science}

\usepackage{fancyhdr}

\begin{document}

\begin{frontmatter}

\title{Exponential Slowdown for Larger Populations: The $(\mu+1)$-EA on Monotone Functions \tnoteref{conference}} 
\tnotetext[conference]{© 2021. This manuscript version is made available under the CC-BY-NC-ND 4.0 license \url{http://creativecommons.org/licenses/by-nc-nd/4.0/}. The formal publication can be found at \url{https://doi.org/10.1016/j.tcs.2021.03.025}.}

\author{Johannes Lengler\corref{1}\fnref{1}}
\ead{johannes.lengler@inf.ethz.ch}
\cortext[1]{Corresponding author}

\author{Xun Zou\corref{1}\fnref{1,2}}
\ead{xun.zou@inf.ethz.ch}

\address{ETH Z{\"u}rich, Z{\"u}rich, Switzerland}

\fntext[1]{Both authors contributed equally to this research.}

\fntext[2]{The author was supported by the Swiss National Science Foundation {[grant number CRSII5\_173721]}.}




\begin{abstract}
Pseudo-Boolean monotone functions are unimodal functions which are trivial to optimize for some hillclimbers, but are challenging for a surprising number of evolutionary algorithms. A general trend is that evolutionary algorithms are efficient if parameters like the mutation rate are set conservatively, but may need exponential time otherwise. In particular, it was known that the $(1+1)$-EA and the $(1+\lambda)$-EA can optimize every monotone function in pseudolinear time if the mutation rate is $c/n$ for some $c<1$, but that they need exponential time for some monotone functions for $c>2.2$. The second part of the statement was also known for the $(\mu+1)$-EA.

In this paper we show that the first statement does \emph{not} apply to the $(\mu+1)$-EA. More precisely, we prove that for every constant $c>0$ there is a constant $\mu_0\in\N$ such that the $(\mu+1)$-EA with mutation rate $c/n$ and population size $\mu_0 \leq \mu \leq n$ needs superpolynomial time to optimize some monotone functions. Thus, increasing the population size by just a constant has devastating effects on the performance. This is in stark contrast to many other benchmark functions on which increasing  the population size either increases the performance significantly, or affects performance only mildly. 

The reason why larger populations are harmful lies in the fact that larger populations may temporarily decrease the selective pressure on parts of the population. This allows unfavorable mutations to accumulate in single individuals and their descendants. If the population moves sufficiently fast through the search space, then such unfavorable descendants can become ancestors of future generations, and the bad mutations are preserved. Remarkably, this effect only occurs if the population renews itself sufficiently fast, which can only happen far away from the optimum. This is counter-intuitive since usually optimization becomes harder as we approach the optimum. Previous work missed the effect because it focused on monotone functions that are only deceptive close to the optimum.
\end{abstract}


\end{frontmatter}
\section{Introduction}
\label{sec:Intro}
Population-based \emph{evolutionary algorithms} (EAs) are general-purpose heuristics for optimization. Having a population may be helpful, because it allows for diversity in the algorithm's states. Such diversity may be helpful for escaping local minima, and it is a necessary ingredient for crossover operations as they are used in \emph{genetic algorithms} (GAs). Theoretical and practical analysis of population-based algorithms have indeed mostly found positive or neutral effects, and showed a general trend that larger populations are better~\cite{witt2006runtime}, or at least not worse than a population size of one~\cite{antipov2020tight}. The only (mild) observed negative effect is, intuitively speaking, that maintaining a population of size $\mu$ may slow down the optimization time by a factor of at most $\mu$. Only few, highly artificial examples are known~\cite{richter2008ignoble, witt2008population} in which a $(\mu+1)$-EA or $(\mu+1)$-GA with time budget $\mu t$ performs significantly worse than a $(1+1)$-EA with time budget $t$. In this sense, it is easy to believe that a $(\mu+1)$ algorithm is at least as good as a $(1+1)$ algorithm, except for the runtime increase that comes from each individual only having probability $1/\mu$ per round of creating an offspring. 

Our results challenge this belief, and show that it is highly wrong for some monotone functions. Our main results show that increasing $\mu$ from $1$ to a larger constant  can increase the runtime from quasilinear to exponential.

A monotone\footnote{Following~\cite{lengler2018drift,lengler2018general}, we call them monotone functions, although \emph{strictly monotone functions} would be slightly more accurate.} pseudo-Boolean function is a function $f:\{0,1\}^n \to \R$ such that for every $x,y \in \{0,1\}^n$ with $x\neq y$ and $x_i \geq y_i$ for all $1 \leq i \leq n$ it holds $f(x) > f(y)$. Monotone functions are easy benchmark functions for optimization techniques, since they always have a unique local and global optimum at the all-ones string. Moreover, from every search point there are short, fitness-increasing paths to the optimum, by flipping zero-bits into one-bits. Consequently, there are many algorithms which can easily optimize every monotone function. A particular example is \emph{random local search} (RLS), which is the $(1+1)$ algorithm that flips in each round exactly one bit, uniformly at random. RLS can never increase the distance from the optimum for a monotone function, and it optimizes any such function in time $O(n \log n)$ by a coupon collector argument. Thus monotone functions are regarded as an easy benchmark for evolutionary algorithms. Nevertheless it was shown in~\cite{lengler2018drift,lengler2018general,DoerrJSWZ10,doerr2013mutation} that a surprising number of evolutionary algorithms need exponential time to optimize some monotone functions, especially if they mutate too aggressively, i.e., the mutation parameter $c$ is too large (see Section~\ref{sec:related} for a detailed discussion). However, in all considered cases the algorithms were efficient if the mutation parameter satisfied $c<1$. 

\subsection{Our Results}\label{sec:ourresults}

We show that the $(\mu+1)$-Evolutionary Algorithm, \moea, becomes inefficient even if the mutation strength is smaller than 1. More precisely, we show that for every $c>0$ there is a $\mu_0=\mu_0(c)\in \N$ such that for all $\mu_0\leq \mu \leq n$ there are some monotone functions for which the \moea with mutation rate $c/n$ needs superpolynomial time to find the optimum. If $\mu$ is $O(1)$ then this time is even exponential in $n$. Note that for $0<c\leq 1$, it is known that the \ooea finds the optimum in quasilinear time for any monotone functions ~\cite{doerr2013mutation, jansen2007brittleness, lengler2019does}. Thus when we increase the population size only slightly (from $1$ to $\mu_0$), the optimization time explodes, from quasilinear to exponential.

The monotone functions that are hard to optimize are due to Lengler and Steger~\cite{lengler2018drift}, and were dubbed \hottopic functions in~\cite{lengler2018general}. These functions look locally like linear functions in which all bits have some positive weights. However, in each region of the search space there is a specific subset of bits (the `hot topic'), which have very large weights, while all other bits have only small weights. If an algorithm improves in the hot topic, then it will accept the offspring regardless of whether the other bits deteriorate. In~\cite{lengler2018drift,lengler2018general,lengler2019general} it was shown that an algorithm like the \ooea with mutation rate $c>2.13..$ will mutate too many of these bits outside of the hot topic, and will thus not make progress towards the global optimum.

The key insight of our paper is that for such weighted linear functions with imbalanced weights, populations may also lead to an accumulation of bad mutations, even if the mutation rate is small. Here is the intuition. For a search point $x$, we call the number of one-bits in the hot topic in $x$ the \emph{rank} of $x$. Consider a \moea close to the optimum, and assume for simplicity that all search points in the population $S_0$ have the same rank $i$. At some point one of them will improve in the hot topic by flipping a zero-bit there. Let us call the offspring $x$, and let us assume that its rank is $i+1$. Then $x$ is fitter than all other search points in the population because it has a higher rank. Moreover, every offspring or descendant of $x$ will also be fitter than all the other points in the population, as long as they maintain rank $i+1$. Thus for a while the \moea will accept all (or most) descendants of $x$, and remove search points of rank $i$ from the population. This goes on until some time $t_0$ at which search points of rank $i$ are completely eliminated from the population. Note that at time $t_0$, most descendants $x'$ of $x$ have considerably smaller fitness than $x$, since the algorithm accepts every type of mutation outside of the hot topic, and most mutations are detrimental. If some descendant $x'$ of $x$ creates an offspring $y$ of even higher rank, then $y$ is accepted and the cycle repeats with $y$ instead of $x$. The crucial point is that $y$ is an offspring of $x'$, which has accumulated a lot of bad mutations compared to $x$. So typically, $x'$ is considerably less fit than $x$, but still it passes on its bad genes. 

The above effect needs that the probability of improving in the hot topic has the right order. If the probability is too large (close to one), then $x$ will already spawn an offspring of rank $i+1$ before it has spawned many descendants with the same rank. On the other hand, if the probability is too small then there will be no rank-improving mutations until time $t_0$, and after time $t_0$ the algorithm starts to remove the worst individuals of rank $i+1$ from the population. We remark that this latter regime was already studied in~\cite{lengler2018general}, for the extreme case in which the improvement probability is so small that typically the population of rank $i+1$ collapses into copies of $x$ before a further improvement is made. (In the terminology of~\cite{lengler2018general}, it was the assumption that the parameter $\eps$ of the \hottopic function was sufficiently small.) However, there is a rather large range of improvement probabilities that lead to the aforementioned effect, i.e., they typically yield an offspring $y$ from some inferior search point $x'$ of rank $i+1$.

\subsection{Related Work}
\label{sec:related}

The analysis of EAs on monotone functions started in 2010 by the work of Doerr, Jansen, Sudholt, Winzen and Zarges~\cite{DoerrJSWZ10,doerr2013mutation}. Their contribution was twofold: firstly, they showed that the \ooea, which flips each bit independently with static mutation rate $c/n$, needs time $O(n \log n)$ on all monotone functions if the mutation parameter $c$ is a constant strictly smaller than one. This result was already implicit in~\cite{jansen2007brittleness}.

On the other hand, it was also shown in~\cite{DoerrJSWZ10,doerr2013mutation} that for large mutation rates, $c>16$, there are monotone functions for which the \ooea needs exponential time. The construction of hard monotone functions in~\cite{DoerrJSWZ10,doerr2013mutation} was later simplified by Lengler and Steger~\cite{lengler2018drift}, who improved the range for $c$ from $c>16$ to $c > c_0 = 2.13..$. Their construction was later called \hottopic functions in~\cite{lengler2018general}, and it will also be the basis for the results in this paper. 

For a long time, it was an open question whether $c=1$ is a threshold at which the runtime switches from polynomial to exponential. On the presumed threshold $c=1$, a bound of~$O(n^{3/2})$ was known due to Jansen~\cite{jansen2007brittleness}, but it was unclear whether the runtime is quasilinear. Finally, Lengler, Martinsson and Steger~\cite{lengler2019does} could show that $c=1$ is not a threshold, showing by an information compression argument an $O(n\log^2 n)$ bound for all $c \in [1,1+\eps]$ for some $\eps >0$. 

Recently, the limits of our understanding of monotone functions were pushed significantly by Lengler~\cite{lengler2018general,lengler2019general}, who analyzed monotone functions for a manifold of other evolutionary and genetic algorithms. In particular, he analyzed the algorithms on \hottopic functions, and found sharp thresholds in the parameters, such that on one side of the threshold the runtime on \hottopic was $O(n \log n)$, while on the other side of the threshold it was exponential. These algorithms include the \ooea, the \olea, the \moea, for which the threshold condition was $c<c_0$, where $c_0 = 2.13..$, and it further included the \ollga, and the so-called `fast \ooea' and `fast \olea'.\footnote{The so-called ``fast'' versions draw the parameter $c$ randomly in each iteration from a heavy-tailed distribution. This avoids that the probability of flipping $k$ bits drops exponentially in $k$~\cite{doerr2017fast}.} Surprisingly, for the genetic algorithms \moga and the `fast \moga', any parameter range leads to runtime $O(n \log n)$ on \hottopic if the population size $\mu$ is large enough, showing that crossover is strongly beneficial in these cases. 

For some of the algorithms, Lengler in~\cite{lengler2018general,lengler2019general} also complemented the results on \hottopic functions by statements asserting that for less aggressive choices of the parameters the algorithms optimize \emph{every} monotone function efficiently. For example, he proved that for mutation parameter $c<1$ and for every constant $\lambda \in \N$, with high probability the \olea optimizes every monotone function in $O(n \log n)$ steps. Analogous statements were proven for the `fast \ooea' and `fast \olea', and for the \ollga, but the condition $c<1$ needs to be replaced by analogous conditions on the parameters of the respective algorithms. Moreover, in the case of the `fast \olea', the result was only proven if the algorithm starts sufficiently close to the optimum. Lengler did not prove any results for general monotone functions for the population-based algorithms \moea and \moga, and for their `fast' counterparts. Our result shows that at least for the \moea, this gap had a good reason. As mentioned before, we will show that for every (constant) mutation parameter $c>0$, there are monotone functions on which the \moea needs superpolynomial time if the population size $\mu$ is larger than some constant $\mu_0 = \mu_0(c)$. It also shows that the \moea and the \olea behave completely differently on the class of monotone functions, since the \olea is efficient for all constant $\lambda$ whenever $c<1$.

Surprisingly, our instance of a hard monotone function is again a \hottopic function. This may appear contradictory to the result in~\cite{lengler2018general,lengler2019general} that the \moea is efficient on \hottopic functions if $c<c_0$. The reason why there is no contradiction is that all the results in~\cite{lengler2018general,lengler2019general} on \hottopic come with an important catch. The \hottopic functions come with several parameters, and we will give the formal definition and a more detailed discussion in Section~\ref{sec:hottopic}. For now it suffices to know that one of the parameters, $\eps$, essentially determines how close the algorithm needs to come to the optimum before the fitness function starts switching between different hot topics. In~\cite{lengler2018general,lengler2019general}, only small values of $\eps$ were considered. More precisely, it was shown that for every $\mu\in\N$ there is an $\eps_0>0$ such that the results for the \moea hold for all \hottopic functions with parameter $\eps \leq \eps_0$, and there were similar restrictions for other parameters of the \hottopic function. In a nutshell, \emph{the effect of switching hot topics was only studied close to the optimum}. Arguably, this was a natural approach since usually the hardest region for optimization is close to the optimum. In this paper, we consider \hottopic functions in a different parameter regime: we study relatively large values of the parameter $\eps$, which is a regime of the \hottopic functions in which the action happens far away from the optimum. \emph{Consequently, the results from~\cite{lengler2018general,lengler2019general} on the \moea on \hottopic do not carry over to the version of \hottopic functions that we consider in this paper.} We stress this point to resolve the apparent contradiction between our results and the results in~\cite{lengler2018general,lengler2019general}.

The above discussion also shows a rather uncommon phenomenon. Consider a small mutation parameter, e.g., $c=1/2$. Our results show that the \moea fails to make progress if the \hottopic function starts switching hot topics far away from the optimum. On the other hand, by the results in~\cite{lengler2018general}, the \moea is not deceived if the \hottopic function starts switching hot topics close to the optimum. Thus, we have found an example where optimization close to the optimum is easier than optimization far away from the optimum, quite the opposite of the usual behavior of algorithms. This strange effect occurs because the problem of the \moea arises from having a non-trivial population. However, close to the optimum, progress is so hard that the population tends to degenerate into multiple copies of a single search point, which effectively decreases the population size to one and thus eliminates the problem (see also the discussion in Section~\ref{sec:ourresults} above). \smallskip

Most other work on population-based algorithms has shown benefits of larger population sizes, especially when crossover is used~\cite{kotzing2018destructiveness,friedrich2015benefit,qian2013analysis,kotzing2011crossover}. Without crossover, the effect is often rather small~\cite{antipov2020tight}. The only exception in which a population has theoretically been proven to be severely disadvantageous is on Ignoble Trails. This rather specific function has been carefully designed to lead into a trap for crossover operators~\cite{richter2008ignoble}, and it is deceptive for $\mu=2$ if crossover is used, but not for $\mu =1$. Arguably, the \hottopic functions are also rather artificial, although they were not specifically designed to be deceptive for populations. However, regarding the larger and more natural framework of monotone functions, our results imply that a \moea with mutation parameter $c=1$ does not optimize all monotone functions efficiently if $\mu$ is too large, while the corresponding \ooea is efficient. 

Moreover, Lengler and Schaller pointed out an interesting connection between \hottopic functions and a dynamic optimization problem in~\cite{lengler2018noisy}, which is arguably more natural. In that paper, the algorithm should optimize a linear function with positive weights, but the weights of the objective function are re-drawn each round (independently and identically distributed). This setting is similar to monotone functions, since a one-bit is always preferable over a zero-bit, and the all-one string is always the global optimum. However, the weight of each bit changes from round to round, which somewhat resembles that the \hottopic function switches between different hot topics as the algorithm progresses. In~\cite{lengler2018noisy} the \ooea was studied, and the behavior in the dynamic setting is very similar to the behavior on \hottopic functions. It remains open whether the effects observed in our paper carry over to this dynamic setting.

\section{Preliminaries and Definitions}
\label{sec:definitions}

\subsection{Notation}
\label{sec:notation}

Throughout the paper we will assume that $f : \{0,1\}^n\to \R$ is a monotone function, i.e., for every $x,y \in \{0,1\}^n$ with $x\neq y$ and such that $x_i \geq y_i$ for all $1\leq i \leq n$ it holds $f(x) > f(y)$. We will consider algorithms that try to maximize $f$, and we will mostly focus on the \emph{runtime} of an algorithm, which we define as the number of function evaluations before the first evaluation of the global maximum of $f$.


For $n\in \N$, we denote $[n] := \{1,\ldots,n\}$. For a search point $x$, we write $\OM(x)$ for the \onemax-value of $x$, i.e., the number of one-bits in $x$. For $x\in \{0,1\}^n$ and $\emptyset \neq I \subseteq [n]$, we denote by $d(I,x) := |\{i\in I \mid x_i=0\}|/|I|$ the \emph{density} of zero-bits in $I$. In particular, $d([n],x) = 1-\OM(x)/n$. Landau notation like $O(n), o(n), \ldots$ is with respect to $n\to \infty$. An event $\mathcal E = \mathcal E(n)$ holds \emph{with high probability} or \emph{whp} if $\Pr[\mathcal E(n)] \to 1$ for $n\to\infty$. A function $f:\N\to\R$ grows \emph{stretched-exponentially} if there is $\delta >0$ such that $f(x) = \exp\{\Omega(n^\delta)\}$, and it grows \emph{quasilinearly} if there is $C>0$ such that $f(x) = O(n\log^C n)$.


Throughout the paper we will use $n$ for the dimension of the search space, $\mu$ for the population size, and $c$ for the mutation parameter. We will always assume that the mutation parameter $c$ is a constant independent of $n$, but the population size $\mu = \mu(n)$ may depend on $n$. 

\subsection{Algorithm}
\label{sec:algorithms}

We will consider the \moea with population size $\mu \in \N$ and mutation parameter $c>0$ for maximizing a pseudo-boolean fitness function $f : \{0,1\}^n\to \R$. This algorithm maintains a population of $\mu$ search points. In each round, it picks one of these search points uniformly at random, the \emph{parent} $x^{t}$ for this round. From this parent it creates an \emph{offspring} $y^{t}$ by flipping each bit of $x^t$ independently with probability $c/n$, and adds it to the population. From the $\mu+1$ search points, it then discards the one with lowest fitness from the population, breaking ties randomly \footnote{We break ties randomly for simplicity. Other selection schemes may give preference to offspring, or generally to more recent search points in case of ties. However, the tie-breaking scheme does not have an impact on our analysis.}.

\begin{algorithm2e}
 \textbf{Initialization:} \\
 \Indp
 $S_0 \assign \emptyset$\;
 \For{$i=1,\ldots,\mu$}{
Sample $x^{(0,i)}$ uniformly at random from $\{0,1\}^n$\;
 $S_0 \assign S_0 \cup \{ x^{(0,i)}\}$\;
 }
 \Indm
 \textbf{Optimization:}	\\
 \Indp
 \For{$t=1,2,3,\ldots$}{
                 \textbf{Mutation:} \\
		Choose $x^{t}\in S_{t-1}$ uniformly at random\; 
		\label{line:mutation} Create $y^{t}$ by flipping each bit in $x^{t}$ independently with probability $c/n$\;
\textbf{Selection:}\\
 Set $S_{t} \assign S_{t-1} \cup \{y^{t}\}$\;

  	\label{line:selection} Select $x \in \arg\min \{f(x)\mid x\in S_t\}$ (break ties randomly) and update $S_{t} \assign S_{t} \setminus \{x\}$\;

	 }
 \caption{The $(\mu+1)$-EA with mutation parameter $c$ for maximizing an unknown fitness function $f:\{0,1\}^n \rightarrow \R$. The population $S$ is a multiset, i.e., it may contain some search points several times.}
 \Indm
\label{alg:mulambda}
\end{algorithm2e}

\subsection{HotTopic Functions}\label{sec:hottopic}
In this section we give the construction of hard monotone functions by Lengler and Steger~\cite{lengler2018drift}, following closely the exposition in~\cite{lengler2018general}. The functions come with five parameters $n\in \N$, $0<\beta < \alpha <1$, $0< \eps <1$, and $L\in\N$, and they are given by a randomized construction. Following~\cite{lengler2018general}, we call the corresponding function $\hottopic_{n,\alpha,\beta,\eps,L} = \HT_{n,\alpha,\beta,\eps,L} = \HT$. 

For $1 \leq i \leq L$ we choose sets $A_i \subseteq [n]$ of size $\alpha n$ independently and uniformly at random, and we choose subsets $B_i\subseteq A_i$ of size $\beta n$ uniformly at random. We define the {\em level} $\ell(x)$ of a search point $x\in\{0,1\}^n$ by 
\begin{equation}\label{eq:level}
\ell(x) := \max \left\{ \ell' \in [L] : d(B_{\ell'}, x) \le \eps \right\},
\end{equation}
where we set $\ell(x)=0$, if no such $\ell'$ exists. Then we define $f: \{0,1\}^n \to \R$ as follows:
\begin{equation}\label{eq:hottopic}
\HT(x) :=\ell(x) \cdot n^{2} +  \sum_{\mathclap{i\in A_{\ell(x)+1}}}x_i\cdot n +  \sum_{\mathclap{i \in R_{\ell(x)+1}}} x_i, 
\end{equation}
where $R_{\ell(x)+1} := [n]\setminus A_{\ell(x)+1}$, and where we set $A_{L+1} := B_{L+1} := \emptyset$. One easily checks that this function is monotone~\cite{lengler2018general}. 

So the set $A_{\ell+1}$ defines the hot topic while the algorithm is at level $\ell$, where the level is determined by the sets $B_i$. Following up on the discussion in the introduction, observe that the level $\ell$ increases if the density of zero-bits in $B_{\ell'}$ drops below $\eps$ for some $\ell' > \ell$. From the analysis we will see that with high probability this only happens if the density of zero-bits in $A_{\ell+1}$ and in the whole string is also roughly $\eps$, up to some constant factors. Hence, the parameter $\eps$ determines how far away the algorithm is from the optimum when the level changes. 

Throughout the paper we will assume that $\alpha$ and $\beta$ are independent of $n$, whereas we will choose small constants $\eta,\rho >0$ and set $\eps = \mu^{-1+\eta}$ and $L = \exp\{\rho \eps n/\log^2 \mu\}$, i.e., $\eps$ and $L$ may depend of $n$, since we also allow $\mu$ to depend on $n$.\footnote{In the papers~\cite{lengler2018drift,lengler2018general,lengler2019general} the parameter $L$ was replaced by a constant parameter $\rho$ such that $L = e^{\rho n}$. This had the advantage that their parameters were all independent of $n$, but since our parameters depend on $n$ anyway, it is more convenient to use the parameter $L$. However, both versions are equivalent.}


\subsection{Tools}\label{sec:tools}
To obtain good tail bounds, we often apply Chernoff's inequality. 
\begin{theorem}[Chernoff Bound~\cite{doerr2018probabilistic}]\label{thm:Chernoff}
	Let $Y_1, \ldots, Y_m$ be independent random variables (not necessarily i.i.d.) that take values in $[0,1]$. Let $S := \sum_{i=1}^m Y_i$, then for all $0 \leq \delta \leq 1$,
	\[
	\Pr[S \leq (1-\delta)\E[S]] \leq e^{-\delta^2\E[S]/2}
	\]
	and for all $\delta \geq 0$,
	\[
	\Pr[S \geq (1+\delta)\E[S]] \leq e^{-\min\{\delta^2,\delta\}\E[S]/3}.
	\]
	Finally, for all $k \ge 2 e \E[S]$,
	\[
	\Pr[S \ge k] \le 2^{-k }.
	\]
\end{theorem}

In addition, we will need the following theorem to bound the sum of geometrically distributed random variables.

\begin{theorem}[Theorem 1 in \cite{witt2014fitness}]
\label{thm:geometry}
	Let $Y_j$, $1\le j \le m$, be independent random variables following the geometric distribution with success probability $p_j$, and let $S := \sum_{j=1}^m Y_j$. If $\sum_{j=1}^m p_j^{-2} \le s < \infty$ then for any $\delta>0$,
	\begin{equation*}
		\Pr[S \le \E[S] - \delta] \le \exp\Big(-\frac{\delta^2}{2s}\Big).
	\end{equation*}
	For $h:=\min\{p_j \mid j \in [m]\}$,
	\begin{equation*}
		\Pr[S \ge \E[S] + \delta] \le \exp\Big(-\frac{\delta}{4}\min \Big\{ \frac{\delta}{s}, h \Big\} \Big).
	\end{equation*}
\end{theorem}

The following lemma estimates useful probabilities, e.g. the probability to improve on the current hot topic.

\begin{lemma}\label{lem:Poisson}
Let $\alpha,c>0$ be constants. Consider a set $A \subseteq [n]$ of size $\alpha n$ where $n$ is large enough, and consider a search point $x\in \{0,1\}^n$.
\begin{enumerate}
	\item The probability that the number of one-bits in $A$ does not decrease after a standard bit mutation with rate $c/n$ on $x$ can be bounded from below by $p_{R} = \etmac/2$.
	
	\item The probability that a standard bit mutation with rate $c/n$ strictly increases the number of one-bits in $A$ has a lower bound $p_L = \eps(x) \alpha c e^{-\alpha c}/2$ and an upper bound $p_U = \eps(x) \alpha c$, where $\eps(x)=d(A, x)$.
	
	\item Let $(1-\eps') \alpha n \le i \le \alpha n$ where $0< \eps' < 1$ and $\eps' n \ge 2 e c$. Let $\rk(x):=|\{j\in A \mid x_j=1\}|$ and let $y$ be an offspring of $x$. If $\rk(x) < i$, then at least one of the following inequalities holds.
		\begin{align*}
			\Pr[\rk(y) \geq i] \leq 2^{-\eps' \alpha n} \qquad \text{or} \qquad \frac{\Pr[\rk(y) \ge i+1]}{\Pr[\rk(y) \ge i]} \le 2\eps'\alpha c.
		\end{align*}
\end{enumerate}
\end{lemma}

\begin{proof}[Proof of Lemma~\ref{lem:Poisson}]
We show the statements one by one.
\begin{enumerate}
	\item One way of creating an offspring with the same number of one-bits in $A$ is to flip no bits at all in $A$. This probability is $(1-c/n)^{\alpha n} = \etmac - O(1/n) \ge \etmac/2$ when $n$ is large enough. 
	
	\item We observe that the probability we consider is at least  
		\begin{align*}
			 \Pr[\text{flip $1$ zero-bit and $0$ one-bits in $A$}] 
			&= \eps(x) \alpha n \cdot \frac{c}{n} \Big(1-\frac{c}{n}\Big)^{\alpha n - 1}\\
			&  = \eps(x) \alpha c \Big(\etmac - O\Big(\frac{1}{n}\Big)\Big) \\
			&\ge \frac{1}{2} \eps(x) \alpha c \etmac.
		\end{align*}
		And it is at most
		\begin{align*}
			\Pr[\text{flip at least $1$ zero-bit}] \le \sum_{i=1}^{\eps(x) \alpha n} \Pr[\text{flip the $i$-th zero-bit}] = \eps(x) \alpha c,
		\end{align*}
		where the second inequality follows from a union bound over all zero-bits in $A$.
		
	\item Assume first that $\rk(x) < (1-2\eps') \alpha n$. Then for $\rk(y)\geq i$, at least $\eps' \alpha n$ zero-bits must be flipped in one mutation. The expected number of flipped zero-bits is at most $\alpha n \cdot c / n = \alpha c$, so that happens with probability $2^{-\eps' \alpha n}$ by the Chernoff bound. So let us consider the other case, $\rk(x) \geq (1-2\eps') \alpha n$. Let $P$ be a permutation on the $\alpha n$ bits in $A$ such that $P(j) < P(j')$ for all $x_j = 1$ and $x_{j'} = 0$. Consider mutating the bits in $x$ in the permuted order, and we track the number $G:= G_0 - G_1$ during that process, where $G_0$ ($G_1$) is the number of flipped zero-bits (one-bits). Clearly, $G$ will be decreasing while we are at the one-bits and increasing afterwards. Then $\rk(y) \ge i$ if and only if $G \ge i -\rk(x)$ after flipping some zero-bit $j$, and $\rk(y) \ge i+1$ if and only if at least one more zero-bit is flipped after bit $j$. The number of remaining zero-bits is at most $\alpha n - \rk(x) - 1 < 2 \eps' \alpha n$, so the probability of flipping at least one remaining zero-bit is at most $2 \eps' \alpha c$ by a union bound. Therefore, 
	\begin{align*}
		\Pr[\rk(y) \ge i+1] &\le 2\eps'\alpha c \cdot \Pr[\text{$G\ge i-\rk(x)$ at some zero-bit $j$}] \\
		&= 2 \eps' \alpha c \cdot \Pr[\rk(y) \ge i]. \qedhere
	\end{align*} 
\end{enumerate}
\end{proof}
 
 We will use the following two theorems to bound the running time of the \moea. The first one states that a sequence of random variables whose differences are small with exponentially decaying tail bound are \emph{sub-Gaussian.}\footnote{The reader can take the concept of being sub-Gaussian as a black box. Theorem~\ref{thm:subgaussian} asserts that exponential tail bounds guarantee the property, Theorem~\ref{thm:bound} describes the consequences. For completeness, we also give the definition: a sequence of random variables $(Y_i)_{i\ge0}$ is $(c, \delta)$-sub-Gaussian if and only if
$
 	\E[\exp(z(Y_{i+1}-Y_i)) \mid Y_0, \ldots, Y_i] \le \exp(z^2 c / 2)
$
 holds for all $i\ge 0$ and $z \in [0, \delta]$.}.
  \begin{theorem}[Timo K{\"o}tzing, Theorem 10  in~\cite{kotzing2016concentration}]\label{thm:subgaussian}
	Let $(Y_i)_{i \ge 0}$ be a supermartingale such that there are $c'>0$ and $\delta'$ with $0<\delta'<1$ and, for all $i\ge 0$ and for all $y\ge 0$,
	\[
	\Pr[|Y_{i+1}-Y_i|\ge y \mid Y_0,\ldots,Y_i] \le c' (1+\delta')^{-y}.
	\]
	Then $(Y_i)_{i\ge0}$ is $(128c'\delta'^{-3},\delta'/4)$-sub-Gaussian.
  \end{theorem}
The other theorem bounds first hitting times of sub-Gaussian supermartingales.
  \begin{theorem}[Timo K{\"o}tzing, Theorem 12  in~\cite{kotzing2016concentration}]\label{thm:bound}
	Let  $(Y_i)_{i\ge 0}$ be a sequence of random variables and let $r\in\R$. If, for all $i \ge 0$,
	\[
	\E[Y_{i+1}-Y_i \mid Y_0,\ldots,Y_i] \le r,
	\]
	then $(Y_i-r i)_{i\ge0}$ is a supermartingale. If further $(Y_i-r i)_{i\ge0}$ is $(c'',\delta'')$-sub-Gaussian, then, for all $i\ge0$ and all $y>0$,
	\[
	\Pr\Big[\max_{0\le j\le i}(Y_j-Y_0) \ge r i+y\Big] \le \exp\Big(-\frac{y}{2}\min\Big(\delta'',\frac{y}{c'' i}\Big)\Big).
	\]
  \end{theorem}

\section{Formal Statement of the Result}

The main result of this paper is the following.

\begin{theorem}\label{thm:main}
For every constant $c>0$ and $0<\beta<\alpha<1$ there exist constants $\mu_0 = \mu_0(c) \in \N$ and $\eta, \rho>0$ such that the following holds for all $\mu_0 \leq \mu \leq n$ where $n$ is sufficiently large. Consider the \moea with population size $\mu$ and mutation rate $c/n$ on the $n$-bit \hottopic function $\HT_{n,\alpha,\beta,\eps,L}$, where $\eps = \mu^{-1+\eta}$ and $L= \lfloor\exp\{\rho\eps n/\log^2 \mu\}\rfloor$. Then with high probability the \moea visits every level of the \HT function at least once. In particular, it needs at least $L$ steps to find the optimum, with high probability and in expectation. 

That is, if $\mu \geq\mu_0$ is a constant (independent of $n$) then with high probability the optimization time is exponential. 
\end{theorem}

We remark that the requirement $\mu \leq n$ is not tight, and we conjecture that the runtime is always superpolynomial for $\mu \geq \mu_0$, also for much larger values of $\mu$. However, we did not undertake big efforts to extend the range of $\mu$ since we do not feel that it adds much to the statement. For larger values of $\mu$, e.g., $\mu = n^2$, our proof does not go through unmodified. With our definition of $\eps = \mu^{-1+\eta}$, we only get error probabilities of the form $\exp\{-\Omega(\eps n/\log^2 \mu)\}$, which are not $o(1)$ if e.g. $\mu = n^2$. Hence we would need to choose larger values of $\eps$, and then we lose a very convenient property, namely that for every fixed $i$, with high probability no individual of rank at most $i-1$ creates an individual of rank at least $i+1$. To avoid these complications, we only consider $\mu \leq n$.

\section{Proof Overview}

The next three sections are devoted to proving Theorem~\ref{thm:main}. The key ingredient is to analyze the drift of the density $d([n],x)$ for search points $x$ which have roughly density $\eps$. We start by giving an informal overview, and by discussing similarities and differences to the situation in ~\cite{lengler2018drift} and~\cite{lengler2018general}.

We will analyze the algorithm in the regime where the fittest search point $x^*$ in the population satisfies 
\begin{align}\label{eq:conditiondensity}
d(A_{\ell+1}, x^*) \in [\eps/2, 2\eps] \quad\text{and}\quad d(R_{\ell+1}, x^*) \in [\eps/2, 2\eps],
\end{align}
where $\ell=\ell(x^*)$ is the current level and $\eps = \mu^{-1+\eta}$ is the parameter of the \hottopic function. It will turn out that for large $\mu$, the algorithm already needs stretched-exponential time to escape this situation. 

The main idea is similar to~\cite{lengler2018drift,lengler2018general}, in which the \ooea and other algorithms were analyzed. We first sketch the main argument for the \ooea, and explain afterwards which parts must be replaced by new arguments. The crucial ingredient is that while the density $d(A_{\ell+1}, x)$ of zero-bits on the hot topic decreases from $2\eps$ to $\eps$, the total density $d([n], x)$ has a positive drift, i.e., a drift away from the optimum. Moreover, the probability to change $k$ bits in one step has a tail that decays exponentially with $k$. Therefore, it was shown that with high probability $d([n],x)$ stays above $\eps+\gamma$ for an exponential number of steps, where $\gamma$ is a small constant. Then it was argued that as long as $d([n],x)$ stays bounded away from $\eps$, it is exponentially unlikely that the level ever increases by more than one. Since there are an exponential number of levels, this implies an exponential runtime.

The analysis of \moea and \moga for constant $\mu$ in~\cite{lengler2018general} was obtained by reducing it to the analysis of a related $(1+1)$ algorithm. This was possible since the choice of parameters in~\cite{lengler2018general} (choosing the parameter $\eps = \eps(\mu)$ sufficiently small) made the algorithm operate close to the optimum. In this range, there are only few zero-bits, and thus it is rather unlikely that a mutation improves the fitness. On the other hand, there is always a constant probability (if $\mu$ is constant) to create a copy of the fittest individual. In such a situation, the population degenerates frequently into a collection of copies of a single search point. Thus, the population-based algorithms behave similarly to a $(1+1)$ algorithm. This $(1+1)$ algorithm has essentially the same mutation parameter as the \moea, while for the \moga it has a much smaller mutation parameter (less than one), which is the reason why the \moga is efficient on all \hottopic instances with small parameter $\eps$. For us, the situation is more complex since we consider larger values of $\eps$. As a consequence, it is easier to find a search point with better fitness, and the population does not collapse. Hence, it is not possible to represent the population by a single point.

Instead, we proceed as follows. Fix a fitness level $\ell$, and consider the auxiliary fitness function 
\begin{align}\label{eq:definition_of_fell}
f_\ell(x) := n\sum_{\mathclap{j \in A_{\ell+1}}} x_j + \sum_{\mathclap{j \in R_{\ell+1}}} x_j.
\end{align} 
We will first study the behavior of the \moea on $f_\ell$. Considering this fitness function is essentially the same as assuming that the level remains the same. We will see in the end that this assumption is justified, by the same arguments as in~\cite{lengler2018drift,lengler2018general}. For a search point $x$, we define the \emph{rank} $\rk(x) := \left|\{j \in A_{\ell+1} \mid x_j=1\}\right|$ of $x$ as the number of correct bits in the current hot topic. Note that by construction of $f_\ell$, a search point with higher rank is always fitter than a search point with smaller rank. 

Now we define $\mathcal X_i$ to be the set of search points of rank $i$ that are visited by the \moea, and we define $\best i$ to be the \onemax-value (the number of one-bits) of the last search point in $\mathcal X_i$ that the algorithm deletes from its population. Note that due to elitist selection, this search point is also (one of) the fittest search point(s) in $\mathcal X_i$ that the algorithm ever visits, and hence it has the largest \onemax-value among all search points in $\mathcal X_i$ that the algorithm ever visits. Then our goal is to show that $\E[\best{i+1} - \best{i}] = -\Omega(1)$, under the assumption that the population satisfies~\eqref{eq:conditiondensity}, i.e., that the density of the fittest search point is close to $\eps$. This assumption can be justified by a coupling argument as in~\cite{lengler2018drift,lengler2018general}. Computing the drift of $\best i$ is the heart of our proof, and the main technical contribution of this paper. In fact, to simplify the analysis we only prove the slightly weaker statement that $\E[\best{i+K} - \best{i}] = -\Omega(1)$ for a suitable constant $K$, which is equally suited. Once we have established this negative drift, the remainder of the proof as in~\cite{lengler2018drift,lengler2018general} carries over almost unchanged. \medskip

To estimate the drift $\Delta := \E[\best{i+K} - \best{i}]$, we will assume for this exposition that $\mu = \omega(1)$, so that we may use $O$-notation. (In the formal proof we will use the weaker assumption $\mu \geq \mu_0$ for a sufficiently large constant $\mu_0 = \mu_0(c)$.) 
We distinguish between \emph{good} and \emph{bad} events. Good events will represent the typical situation; they will occur with high probability, and if they occur $K$ times in a row, then it will deterministically  follow that $\best{i+K} - \best{i} \leq -\log \mu$. On the other hand, bad events may lead to a positive difference, but they are unlikely and thus they contribute only a lower order term to the drift. We will discriminate two types of bad events. Firstly, we will show that the probability $\Pr[\best{i+K} - \best{i} > \lambda \log \mu]$ drops exponentially in $\lambda$. This implies that the events in which $\best{i+K} - \best{i} > \log^{2} \mu$ contribute at most a term $o(1)$ to the drift. Hence, we can restrict ourselves to the case that $\best{i+K} - \best{i} \leq \log^{2} \mu$. Now assume that we have any event of probability $o(\log^{-2} \mu)$. In the case $\best{i+K} - \best{i} \leq \log^{2} \mu$, this event can contribute at most a $o(1)$ term to the drift. Hence, we may declare any such event as a bad event, and conclude that all bad events together only contribute a $o(1)$ term to the drift.

As we have argued, we may neglect any event with probability $o(\log^{-2} \mu)$. This is a rather large error probability, which allows us to dub many events as `bad', and to use rather coarse estimates on the error probability. 
We conclude this overview by describing how a good event, and thus a typical situation, looks like. In what follows, all claims hold with probability at least $1-o(\log^{-2} \mu)$. \medskip

Let us call $t_i$ the first round in which an individual of rank at least $i$ is created, and $T_i$ the round in which the last individual of rank at most $i$ is eliminated. Then typically $T_i-t_i = O(\mu \log \mu) \cap \Omega(\mu)$. Let $\left|X_i\right| = \left|X_i(t)\right|$ denote the number of search points in the population of rank $i$ at time $t$. We want to study the \emph{family forest} $F_i$ of $X_{\geq i}$, which is closely related to the family trees and family graphs that have been used in other work on population-based EAs, e.g.~\cite{witt2006runtime,antipov2020tight,lehre2012impact,sudholt2009impact}. The vertices of this forest are all individuals of rank at least $i$ that are ever included into the population. A vertex is called a \emph{root} if its parent has rank less than $i$. Otherwise, the forest structure reflects the creation of the search points, i.e., vertex $u$ is a child of vertex $v$ if the individual $u$ was created by a mutation of $v$.

As $X_i$ grows, eventually the first few search points of rank $i+1$ are created, and form the first roots of the family forest. Then the forest starts growing, both because new roots may appear and because the vertices in the forest may create offspring. At some point we have $|X_{i+1}| = \mu^{\delta}$ for some (suitably small) $\delta >0$. At this point, we still have typically $|X_i| = O(\mu^{\delta}/\eps) = O(\mu^{1+\delta-\eta}) = o(\mu)$, where the latter holds if $\delta$ is small enough. Moreover, at this point there are no search points of rank strictly larger than $i+1$. The sets $X_i$ and $X_{i+1}$ both continue to grow with roughly the same speed until the search points of rank at most $i-1$ are eliminated from the population. Afterwards, the search points of rank $i$ are eliminated from the population, until only search points of rank at least $i+1$ remain. Crucially, up to this point every search point of rank at least $i+1$ is accepted into the population. In other words, there is no selective pressure on the search points of rank $i+1$, and every mutation of a search point of rank $i+1$ enters the family tree, as long as the rank $i+1$ is preserved. Therefore, we can contain the family forest $F_{i+1}$ of rank $i+1$ up to this point in a random forests $F'$ which is obtained by certain forest growth processes in which no vertex is ever eliminated and all vertices continue to spawn offspring with a fixed rate. 

We want to understand the set of individuals in $X_{i+1}$ that spawn offspring in $X_{i+2}$, and thus spawn the roots for the family forest $F_{i+2}$. As before we can argue that no individuals of rank at least $i+2$ are created before the family forest of rank $i+1$ reaches size $\mu^{\delta}$. 
Moreover, we can show that the time $T_{i+1}$ at which all individuals of rank $i+1$ are eliminated from the population satisfies $T_{i+1} - t_{i+1} \leq C\mu\log\mu$ for a suitable constant $C>0$. Hence, $F_{i+1}$ is bounded from above by the random forest $F'$ at time $t_{i+1} +C\mu\log\mu$. This forest is only polynomially large in $\mu$.

\begin{figure}[h]
 \centering
  \includegraphics[width=1.0\textwidth]{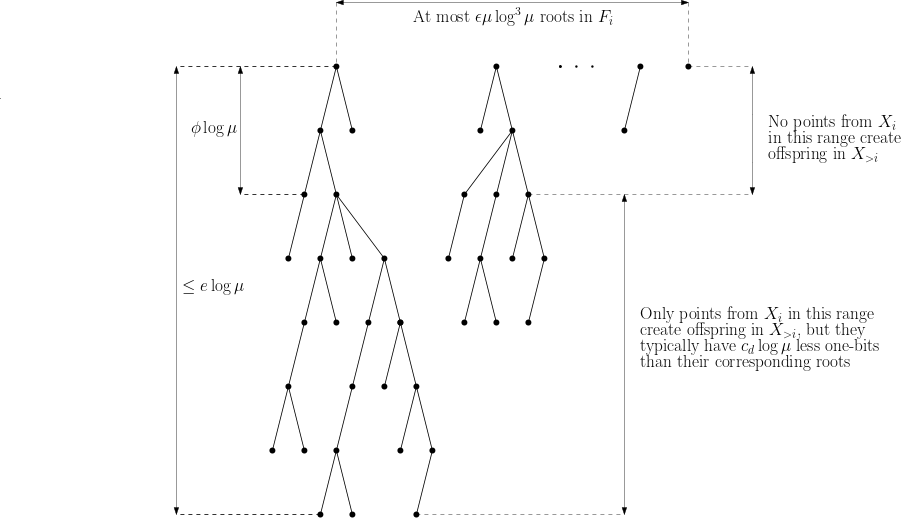}
 \caption{A depiction of the family forest $F_i$, where $\phi$ and $c_d$ are constants to be introduced in Section \ref{sec:typical}. The same picture also applies to its upper bound $F'$. }\label{fig:familyforest}
 \end{figure}

The recursive trees that we use to bound $F_{i+1}$ are well understood, see also Figure~\ref{fig:familyforest}. In particular, it is known that even in $F'$ only a small fraction $\mu^{\delta}$ of the vertices are in depth at most $\phi \log\mu$, where $\delta,\phi >0$ are suitable constants. Since each such vertex creates an offspring of strictly larger rank with probability $\eps/\mu$ per round, the expected number of offspring of rank $i+2$ of these vertices is at most $O(\mu^{\delta}\eps/\mu\cdot(T_{i+1}-t_{i+1}))$. With the right choice of parameters, this is $\mu^{-\Omega(1)}$, and we may conclude that no vertices of depth at most $\phi\log\mu$ create roots of rank $i+2$. On the other hand, since we do not truncate any vertices in the creation of $F'$, they are obtained from their parents by unbiased mutations of $[n]\setminus A_\ell$, and we can show that most (all but at most $\mu^\delta$) vertices of depth at least $\phi \log \mu$ in $F'$ have accumulated $c'\log \mu$ more bad than good bit-flips when compared to their roots, for a suitable $c'>0$. For the $\mu^{\delta}$ exceptional vertices, none of them will create a root of rank $i+2$ in $T_{i+1}-t_{i+1}$ rounds, even if they are in $F_{i+1}$.

To summarize, good events consist of the following four main points. Firstly, no vertex of rank at most $i$ creates an offspring of rank at least $i+2$. Secondly, every vertex in $X_{i+1}$ that creates an offspring in $X_{i+2}$ has at least depth $\phi \log \mu$ in the family forest. Thirdly, every vertex in $X_{i+1}$ of depth at least $\phi \log \mu$ that creates an offspring in $X_{i+2}$ has a \onemax value that is at least $c'\log \mu$ smaller than that of its root. Finally, we also require that no vertex in $F'$ exceeds the \onemax value of its root by more than $C\log\mu$, for some $C>0$. The complete list in the proof contains even more requirements, but these four already imply a decline in $\best i$ if they hold over $K$ consecutive steps. In this case, inductively the \onemax values of all roots in $F_{i+K}$ are at most $\best{i} - K c' \log \mu$. Moreover, $\best{i+K}$ exceeds the \onemax value of the corresponding root in $X_{i+K}$ by at most $C\log\mu$, so we have $\best{i+K} \leq \best{i} - K c' \log \mu + C\log \mu$. Choosing $K$ sufficiently large shows that $\best{i}$ must decrease in these typical situations.

\section{Drift of $\best i$}\label{sec:proofs}

In this main section of the proof, we show that the random variable $\best i$ has negative drift. We will use the same notation as in the proof outline. In particular, $X_i$ denotes the set of all search points of rank $i$ that the algorithm visits, and $\best i$ denotes the \onemax-value of the last search point from $X_i$ that the algorithm keeps in its population. If $X_i$ is empty (which, as we will see, is very unlikely), then we set $\best{i} := \best{i-1}$. Moreover, we define $X_{\ge i} := \bigcup_{i'\ge i} X_{i'}$, and the definition of terms like $X_{> i}$ is analogous. For a given parent individual $x$, we denote by $p_I$ (by $p_R$) the probability that an offspring of $x$ has rank which is strictly larger than (at least as large as) the rank of $x$.

Throughout this section, we fix a level $\ell$ and consider the \moea on the linear function $f_\ell$ defined in~\eqref{eq:definition_of_fell}. In this section, we will study the case that $i \in [(1-2\eps) \alpha n, (1-\eps/2)\alpha n]$, where $\eps = \mu^{-1+\eta}$. Note that this is a weaker form of Condition~\eqref{eq:conditiondensity}, i.e., we consider search points for which the density in $A$ is close to $\eps$.

\subsection{Preliminaries}
\label{subsec:growth}
In this section we first give bounds on the time that the set $X_{\ge i}$ needs to grow from size $1$ to size $\mu^\kappa$, and we will conclude that $X_{\ge i}$ is large at the latter point in time. We start by bounding the time. 
\begin{lemma}
\label{lemma:growth}
	For all $0<\alpha<1$, $c>0$, $0<\eta < \kappa \leq 1$, there exists a constant $\mu_0$ such that the following holds for all $\mu_0 \leq \mu \leq n$. Let $i > (1-2\eps) \alpha n$, where $\eps =\mu^{-1+\eta}$. Consider the \moea with mutation rate $c/n$ on the linear function $f_\ell$. Denote by $T_i^\kappa = T^\kappa$ the number of rounds until $|X_{\geq i}|$ reaches $\mu^\kappa$ after the algorithm visits the first point $x^{i}$ in $X_{\ge i}$. With probability $1-2\mu^{-\Omega(1)}$,
	\begin{align*}
	 \frac{1}{2} (\kappa-\eta)\mlm \le T^\kappa \le 4 \kappa e^{\alpha c} \mlm.
	\end{align*} 
	Moreover, 
	\begin{align*}
		\E[T^\kappa] \le 3 \kappa e^{\alpha c} \mlm.
	\end{align*} 
\end{lemma} 
\begin{proof}
By the definition of $f_{\ell}$, all individuals in $X_{\geq i}$ are fitter than those in $X_{<i}$. So no points in $X_{\geq i}$ will be discarded until $X_{<i}$ becomes empty, and we are interested in the growth of $|X_{\geq i}|$ during this period. 
Let $T_j$ be the time needed for $|X_{\geq i}|$ to grow from $j$ to $j+1$. By definition we have $T^\kappa = \sum_{j=1}^{\mu^\kappa-1} T_j$. Denote by $x^t$ the point selected as parent by the algorithm in round $t$ and denote by $y^t$ its offspring. The probability that both $x^t$ and $y^t$ belong to $X_{\geq i}$ is at least $p_j = j/\mu\cdot p_R$, where $j$ is the size of $X_{\geq i}$ at the beginning of round $t$ and $p_R=e^{-\alpha c}/2$ is defined in Lemma \ref{lem:Poisson}.1.
It is clear that we can dominate $T_j$ by random variable $\bar{T}_j$ that follows a geometric distribution with parameter $p_j$. By Lemma 1.8.8 in \cite{doerr2020probabilistic}, $T^\kappa$ is dominated by $\bar T^\kappa := \sum_{j=1}^{\mu^\kappa-1} \bar T_j$. Next we apply Theorem \ref{thm:geometry} to bound $\bar T^\kappa$ from above. 

The expectation of $\bar T^\kappa$ is
\begin{equation*}
     \E[\bar{T}^\kappa] = \sum_{j=1}^{\mu^\kappa-1} \E[\bar{T}_j] \le 2 e^{\alpha c} \mu \sum_{j=1}^{\mu^\kappa} \frac{1}{j}.
\end{equation*}
For the Harmonic series, we have $\log(m+1) < \sum_{j=1}^m  1/j \le \log m +1$, where $\log$ denotes the natural logarithm. Therefore, for large enough $\mu$,
\begin{equation}
\label{eq:uppert}
   \E[T^\kappa] \le \E[\bar{T}^\kappa] \le 2\etac\mu(\log(\mu^\kappa)+1) \le 3 \kappa \etac \mlm.
\end{equation}
Let $h:=\min\{p_j \mid j=1,\ldots,\mu^\kappa-1\}$, clearly $h = p_1  = \etmac/(2\mu)$. Let $s:=\sum_{j=1}^{\mu^\kappa-1} p_j^{-2}$, we have 
\begin{align*}
    s \le 4 e^{2\alpha c} \mu^2 \sum_{j=1}^{\mu^\kappa} \frac{1}{j^2} 
    \le \frac{2 e^{2 \alpha c} \pi^2}{3} \mu^2,
\end{align*}
where the last step follows from $\sum_{j=1}^\infty 1/j^2 = \pi^2/6$.
Given $h$ and the bound on $s$, by Theorem \ref{thm:geometry} it holds for $\delta = \kappa \etac \mu\log\mu$ that 
\begin{align*}
	\Pr\left[\bar{T}^\kappa \ge \E[\bar{T}^\kappa] + \delta\right] \le e^{-\Omega(\log\mu)} = \mu^{-\Omega(1)}.
\end{align*}
Since $T^\kappa \preceq \bar T^\kappa$, together with equation (\ref{eq:uppert}) we conclude that $T^\kappa \le 4 \kappa e^{\alpha c} \mlm$ with probability $1 - \mu^{-\Omega(1)}$.

We still need a lower bound of $T^\kappa$. Consider the probability that $\Xgi$ gets a new offspring $y^t$ in a round where $|\Xgi| = j$:
\begin{align*}
	\Pr\left[y^t \in X_{\ge i}\right] 
	&= \Pr\left[x^t \not\in \Xgi \land y^t \in \Xgi \right] + \Pr\left[x^t \in \Xgi \land y^t \in \Xgi  \right] \\
	&\le (\mu-j)/\mu \cdot p_U + j/\mu \cdot 1  \le j/\mu + p_U,
\end{align*}
where $p_U$ is defined in Lemma \ref{lem:Poisson}.2. Let $p_j' = j/\mu + p_U$, similarly as for the upper bound on $T^\kappa$, we can subdominate $T^\kappa$ with a random variable $\hat T^\kappa = \sum_{j=1}^{\mu^\kappa-1}\hat T_j$ (Lemma 1.8.8 in \cite{doerr2020probabilistic}), where the $\hat T_j$ are independent and geometrically distributed with parameter $p_j'$, respectively. Then
\begin{align*}
	\E[\hat T^\kappa] &\ge \sum_{j=1}^{\mu^\kappa-1} \frac{1}{p_j'} = \sum_{j=1}^{\mu^\kappa-1} \frac{\mu}{j+\mu p_U}\\
	&\ge \sum_{j=1}^{\mu^\kappa-1} \frac{\mu}{j+\ceil{\mu p_U}} = \sum_{j=1}^{\mu^\kappa-1+\ceil{\mu p_U}} \frac{\mu}{j} - \sum_{j'=1}^{\ceil{\mu p_U}} \frac{\mu}{j'} \\
	&> \mu \log\big(\mu^\kappa+\ceil{\mu p_U}\big) - \mu\big(\log\ceil{\mu p_U} + 1\big).
\end{align*}
 Since $i\ge(1-2\eps)\alpha n$, $P_U=O(\eps) = O(\mu^{-1+\eta})$ for $0 < \eta < \kappa$. So $\ceil{\mu p_U} = O(\mu^\eta)$. Hence,
\begin{align*}
\label{eq:lowert}
	\E[T^\kappa] \ge \E[\hat T^\kappa] \ge \big(1-O\big(\log^{-1}\mu\big)\big) (\kappa - \eta) \mlm.
\end{align*}
Let $s':=\sum_{j=1}^{\mu^\kappa} p_j'^{-2}$. As $p_j' > p_j$, it holds $s' < s$ that. Applying Theorem \ref{thm:geometry} with $s'$ and $\delta' = \eps' \mu \log\mu$, we obtain
\begin{align*}
	\Pr\left[\hat{T}^\kappa \le \E[\hat{T}^\kappa] - \delta'\right] \le e^{-\Omega(\log^2\mu)} = \mu^{-\Omega(1)}.
\end{align*}
Similarly, we have $\hat T^\kappa \preceq T^\kappa$, by picking a sufficiently small $\eps'$ we conclude that
\begin{equation*}
	T^\kappa \ge \frac{1}{2} (\kappa-\eta) \mlm
\end{equation*}
with probability $1 - \mu^{-\Omega(1)}$.
%
\end{proof}

In the following lemma, we give a lower bound on $|X_{\ge i+1}|$ when $X_{\ge i}$ reaches a certain size.
\begin{lemma}
	\label{lemma:improvement}
	Let $\alpha, \kappa \in (0,1)$, $c>0$, $\eta<1$ be constants such that $\kappa > 1-\eta/2$. Consider the \moea with $\mu \leq n$ and mutation rate $c/n$ on the linear function $f_\ell$. Let $\eps = \mu^{-1+\eta}$ and let $i \le (1-\eps/2)\alpha n$. Denote by $Y_{i+1}^\kappa = Y^\kappa$ the size of $X_{\ge i+1}$ when $|X_{\ge i}|$ reaches $\mu^\kappa$. Then with probability $1-\exp\big(-\Omega\big(\mu^{2(\kappa-1)+\eta}\big)\big)$,
	\begin{align*}
	Y^\kappa = \Omega\big(\eps \mu^{2\kappa-1}\big) = \Omega\big(\mu^{2(\kappa-1)+\eta}\big) = \mu^{\Omega(1)}.
	\end{align*}
\end{lemma} 

\begin{proof}

Note that we may assume that $\mu \geq \mu_0$ for a constant $\mu_0$ of our choice, since otherwise the probability may be zero and thus the statement is vacuous. In each round $|X_{\ge i}|$ increases by either 0 or 1, so after $X_{\ge i}$ reaches size $R:=\floor{\mu^\kappa/2}$ there are at least $R$ more rounds until $|X_{\ge i}|=\mu^\kappa$. In each of the remaining $R$ rounds, the probability of a parent $x \in X_{\ge i}$ being selected and its offspring $y$ belonging to $X_{> i}$ is at least
\begin{equation*}
	\Pr[y \in X_{>i}] > \Pr[x\in X_{\ge i} \land y \in X_{>i}] \ge R/\mu \cdot p_L,
\end{equation*}
where $P_L$ is defined in Lemma \ref{lem:Poisson}.
Let $Y_j$ be independent Bernoulli variables with parameters $ R/\mu \cdot p_L$ for $j\in[R]$. Then $Y^\kappa$ dominates the sum of $Y_j$, i.e. $Y^\kappa \succeq \bar Y^\kappa := \sum_{j=1}^{R} Y_j$. It holds that 
\begin{equation*}
	\E[\bar Y^\kappa] = R \cdot R/\mu \cdot p_L = \Theta\big(\eps \mu^{2\kappa-1}\big) =  \Theta\big(\mu^{2(\kappa-1)+\eta}\big).
\end{equation*}
By Chernoff's inequality (Theorem \ref{thm:Chernoff}), we have for any constant $0<\delta<1$,
\begin{equation*}
	\Pr\big[\bar Y^\kappa < (1-\delta)\E[\bar Y^\kappa]\big] \le \exp\big(-\Omega\big(\mu^{2(\kappa-1)+\eta}\big)\big).
\end{equation*}
The claim follows from $Y^\kappa \succeq \bar Y^\kappa$.
\end{proof}

\subsection{Tail Bounds}

In this section, we will give rather loose tail bounds to show that it is unlikely that $\best i$ is much larger than $\best{i-1}$. All constants in this section are independent of $\mu$. This includes all hidden constants in the $O$-notation.

\subsubsection{Tail Bound on the Lifetime of $X_i$}
As before, let $\tbirth i$ be the first round in which an individual of rank at least $i$ is created, and let $\tdie i$ be the round in which the last individual of rank at most $i$ is eliminated. 

\begin{lemma}\label{lem:findgoodlabel1}
	For all $0<\alpha,\eta<1$, $c>0$, there is a constant $\mu_0\in\N$ such that the following holds for all $\mu_0 \leq \mu\leq n$. Let $i \in [(1-2\eps) \alpha n,(1-\eps/2)\alpha n]$, where $\eps =\mu^{-1+\eta}$. Consider the \moea with mutation rate $c/n$ on the linear function $f_\ell$. Then with probability at least $1-\mu^{-\Omega(1)}$, $T_i-t_i \le 8\etac\mlm$. Moreover,
	for all $\beta \geq 1$ and $C = 16\etac$,
\begin{align*}
    \Pr[\tdie i - \tbirth i \geq \beta\cdot C\mu\log \mu]\leq 2^{-\beta}.
\end{align*}
\end{lemma}

\begin{proof}
We first show that $\Pr[\tdie i - \tbirth i \geq C' \mu \log \mu] \leq 1/2$ for a suitable constant $C' >0$. Let $x^{\ge i}$ be the first individual of rank at least $i$ and let $x^{j}$ with rank $j$ be the first individual of rank strictly larger than $i$. We can divide the process from $t_i$ to $T_i$ into two parts. The first part ends when $x^{j}$ is created, and we denote by $t_{j}$ the round when this happens. The second part starts after $t_{j}$ and ends when $X_{> i}$ reaches size $\mu$. Since we are proving an upper bound of the tail, we can consider the second part ends when $X_{\ge j}$ reaches $\mu$ for simplicity.

If $x^{\ge i} = x^j$, then we have $t_j=t_i$, namely the first part does not exist. So for the tail bound of the first part, we may assume that $x^{\ge i} \in X_i$. By Lemma \ref{lemma:growth}, for some $1-\eta/2<\kappa<1$, we have $|X_{\ge i}| \geq \mu^\kappa$ at time $T:=t_i+4\kappa\etac\mlm$. By Lemma \ref{lemma:improvement} we have $|X_{> i}| > 0$ at this point, so $x_j$ must have been created before time $T$. For the second part, we apply Lemma \ref{lemma:growth} again for $X_{\ge j}$. By time $t_j+4\etac\mlm$, $X_{\ge j}$ reaches size $\mu$.

To summarize, we have applied Lemma \ref{lemma:growth} twice and Lemma \ref{lemma:improvement} once. Therefore, with probability at least $1-5\mu^{-\Omega(1)}$, $T_i-t_i \le 8\etac\mlm$. Since $\mu \ge \mu_0$, for large enough $\mu_0$ we obtain $\Pr[\tdie i - \tbirth i \geq C' \mu \log \mu] \leq 1/2$ for $C' = 8\etac$. 

To conclude the proof, we set $C := 2C'$. Then for all integral $\beta' \in \N$ we consider $\beta'$ phases and repeat the same argument. This shows $\Pr[\tdie i - \tbirth i \geq \beta'\cdot C'\mu\log \mu]\leq 2^{-\beta'}$.
Hence, for $C = 16 \etac$ it holds for all $\beta \geq 1$,
\begin{align*}
    \Pr[\tdie i - \tbirth i \geq \beta\cdot C\mu\log \mu]&\leq \Pr[\tdie i - \tbirth i \geq \lceil\beta\rceil C'\mu\log \mu] \leq 2^{-\lceil\beta\rceil} \leq 2^{-\beta}. \qedhere
\end{align*}
\end{proof}

\subsubsection{Family Forests}

From now on we will be mostly working on family forests, so we introduce the definition and several related lemmas here. The main idea is to couple the algorithm with a process that is not subject to selection. This idea has been used before to analyze population-based algorithms~\cite{witt2006runtime,antipov2020tight,lehre2012impact,sudholt2009impact}.

We denote the \emph{family forest} for search points with rank at least $i$ by $F_i$. The vertex set of $F_i$ are the vertices in $X_{\ge i}$ that are (once) in the population, while the roots of the trees are vertices whose parents are in $X_{< i}$. Moreover, any path connecting a root and a vertex in $F_i$ corresponds to a series of mutations that create this vertex. Note that the size of $F_i$ increase over time.

As analysing $F_i$ directly can be complicated, we couple it with a simpler random forest $F'$, which is generated by the following process. In round 0 there is a single root in $F'$. In each subsequent round, each vertex in $F'$ creates a new child with probability $1/\mu$ and a new root is added. Lemma \ref{lemma:coupling} shows that $F_i$ can be coupled to a subgraph in $F'$.

\begin{lemma}
\label{lemma:coupling}
The family forest $F_i$ can be coupled to $F'$ such that $F'$ contains $F_i$ as a subgraph at any round.  
\end{lemma}

\begin{proof}
Throughout the coupling process we maintain that $F_i$ is a subgraph of $F'$. The first point $x^{\ge i}$ that the algorithm visits in $X_{\ge i}$ (in round $t_i$) corresponds to the only root $r_0$ in round 0 in $F'$. In every round $t > t_i$, a point $x^t$ in the current population is selected to create an offspring $y^t$. For each $x \in F_i$, if $x^t = x$ (which happens with probability $1/\mu$ if $x$ is still in the current population, and with probability zero otherwise) then we attach a child to $x$ in $F'$: if $y^t \in X_{\ge i}$ then we attach $y^t$ to $x$ in $F'$, otherwise we attach a dummy child to $x$ in $F'$. In this case, we still associate the offspring with the dummy child, and in our upcoming considerations we will ignore that this search point does belong to $X_{\ge i}$. If $x^t$ is not in $X_{\ge i}$ while $y^t$ is, we add $y^t$ as a new root $r_t$ to $F'$, otherwise we add a new dummy root to $F'$. For every node $x \in F'$ that is a dummy node (that has no corresponding node in $F$) or whose copy in $F$ has been removed from the population, we add another dummy node as its child with probability $1/\mu$. In this way, for each vertex in $F'$ we create a new child with probability $1/\mu$ and a root is added in each round. On the other hand, by construction, $F_i$ is a subgraph of $F'$ at all times.
\end{proof}

Note that the search points associated with the vertices in $F'$ are obtained from the root by mutation only, without any interfering selection step. This makes the process easy to analyze. Such a selection-free mutation process has been analyzed before, e.g.~\cite{lassig2013design}.
In Lemma \ref{lemma:forest} we show several useful properties of $F'$. Due to the coupling from $F_i$ to $F'$, the properties will also hold for $F_i$ as well.

\begin{lemma}
\label{lemma:forest}
$F'$ satisfies the following properties:
\begin{enumerate}
	\item Let $s_t$ denote the number of vertices in $F'$ in round $t$, then $\Pr[s_t \ge S] \le t e^{t/\mu} /S$ for all $S>0$.
	
	\item Let $x$ be a search point that corresponds to a vertex in $F'$ of depth at most $d$ with root $y$. Then for $k \ge 2 e d c$,
	\begin{align*}
	    \Pr[\text{$x$ and $y$ differ in more than $k$ bits}]\leq 2^{-k}.
	\end{align*}
	\item Let $x$ be a search point that corresponds to a vertex in $F'$ of depth larger than $d$ with root $y$. If $n$ is sufficiently large and $\OM(y) \geq (1-8\eps)n$ then
	\begin{align*}
	    \Pr[\text{$x$ has more one-bits than $y$}]\leq 2 e^{-d c / 32}
	\end{align*}
	and
	\begin{align}\label{eq:evenmoreonebits}
		\Pr[\text{$y$ has less than $d c/16$ more one-bits than $x$}] \le 2 e^{-d c / 128}.
	\end{align}
	If $n$ is sufficiently large and $\OM(y) \leq (1-8\eps)n$ then $\Pr[\OM(x) \geq (1-4\eps)n] \leq 2 \cdot 2^{-\eps n}$.
	
	\item Let $s_t^d$ denote the number of vertices of depth $d$ in round $t$ for an arbitrary tree from $F'$. Then
	\begin{align*}
		\E[s_t^d] \le \frac{t^d}{d! \mu^d}.
	\end{align*}
	In particular, for $t=O(\mu\log\mu)$ the depth of the tree is at most $e \log \mu$ with probability $1-\mu^{-\Omega(1)}$. 
	Moreover, if $t \ge 2d\mu$, $\sum_{i=0}^d \E[s_t^i] \le 2 t^d / (d! \mu^d)$.
	
\end{enumerate}
\end{lemma}

\begin{proof}
We prove the statements one by one.
\begin{enumerate}
	\item In $t$ rounds we have added $t$ roots to the forest, and we will give a uniform bound for all of them. So we fix a root and denote by $\sigma_{\tau}$ the number of vertices in this tree in round $\tau$, where $0 \le \tau \le t$. We assume pessimistically that the root is introduced in round $0$. Then we have $\sigma_0 = 1$ and $\E\left[ \sigma_{\tau+1} \mid \sigma_{\tau}\right] = (1+1/\mu) \sigma_{\tau}$ for $0 \le \tau \le t-1$. By linearity of expectation, we have $\E[\sigma_{t}] \leq (1+1/\mu)^{t}$. Since there are $t$ roots, and using that $(1+1/\mu)^\mu \leq e$, we obtain
	\begin{align*}
		\E[s_t] \leq  t \E[\sigma_t] \le t (1+1/\mu)^t \le t e^{t/\mu}.
	\end{align*}
	By Markov's inequality, it holds that
	\begin{align*}
	\Pr[s_t \geq S] \leq \frac{\E[s_t]}{S} \le \frac{t e^{t/\mu}}{S}.
	\end{align*}
	
	\item Let $y_i$ be the $i$-th bit in $y$, the event $y_i \neq x_i$ implies that the $i$-th bit is flipped at least once. Denote by $d' \le d$ the distance between $x$ and $y$. By a union bound
	\begin{align*}
	\Pr[y_i \neq x_i] &\le \Pr[\text{bit $i$ is flipped at least once}] \\
		&\le d' \Pr[\text{bit $i$ is flipped in one mutation}] \le d c / n.
	\end{align*}
	Let $D=|\{i \in [n] \mid y_i \neq x_i\}|$ be the number of bits in which $y$ and $x$ differ. Then its expectation is $\E\left[D\right] \le d c$. Since the bits are modified independently, we can apply Chernoff's inequality (Theorem \ref{thm:Chernoff}) for $k \ge 2 e d c \ge 2 e \E[D]$ , and obtain
	\begin{align*}
	\Pr\left[D \ge k \right] \le 2^{-k}.
	\end{align*}
	
	\item Let the depth of $x$ be $d' \ge d$. First we argue that we may assume $d'\leq n/(16 e c)$. If $d'\geq n/(16 e c)$, then consider just the last $n/(16 e c)$ steps. In these, every bit has a constant probability to be touched exactly once, and a constant probability not to be touched at all. If the number of one-bits before the last $n/(16 e c)$ steps was at least $n/2$, then with probability $1-e^{-\Omega(n)}$, $x$ has at least $8\eps n$ zero-bits as each one-bit has a constant probability of being flipped exactly once in those steps, and if the number of one-bits was at most $n/2$, $x$ has also at least $8\eps n$ zero-bits as each zero-bit has a constant probability of being untouched. In either case, $x$ has more zero-bits than $y$ with sufficiently large probability. So we may assume $d'\leq n/(16 e c)$. 
	
	We then consider the case $\OM(y) \geq (1-8\eps)n$. Let $B_{01}$ be the number of bits flipped from 0 to 1. Then similarly as for Property 2 we bound $\E[B_{01}] $ by
	\begin{align*}
		&\quad \left| \{i \mid y_i = 0\} \right| \cdot \Pr\left[x_i = 1 \mid y_i = 0\right] \\
		&\le \left| \{i \mid y_i = 0\} \right| \cdot \Pr[\text{bit $i$ flipped at least once in $d'$ mutations}] \\
		&\le 8\eps n \cdot(d' c/n) 
		 = 8 \eps c d',
	\end{align*}
	where the second inequality follows from a union bound.
	Similarly, let $B_{10}$ be the number of bits flipped from 1 to 0 in $d'$ mutations, its expectation $\E[B_{10}]$ is 
	\begin{align*}
	&\quad \left| \{i \mid y_i = 1\} \right| \cdot \Pr\left[x_i = 0 \mid y_i = 1\right] \\
	&\ge \left| \{i \mid y_i = 1\} \right| \cdot \Pr[\text{bit $i$ flipped exactly once in $d'$ mutations}] \\
	&\ge \frac{n}{2} \binom{d'}{1} \frac{c}{n} \left(1-\frac{c}{n}\right)^{d'-1} 
	\ge  \frac{d' c}{2} \left(1 - \frac{c}{n} d'\right) 
	\ge \frac{d' c}{4}.
	\end{align*}
	Since all bits contribute independently, we may apply the Chernoff bound. With probability at least $1-e^{-d'c/32}$ each, we have $B_{01} \leq cd'/8$ and $B_{10} \geq cd'/8$. Both inequalities together imply that $\OM(x) \leq \OM(y)$ as desired, and the probability that at least one of the inequalities is violated is at most $2e^{-d'c/32} \leq 2e^{-dc/32}$.
	
	Similarly, the probability that $B_{01}$ ($B_{10}$) overshoot (undershoot) its expectation by more than $d' c/16$ is at most $e^{-d' c/128}$. Therefore, the probability that $B_{01} \ge B_{10} - d' c/16$ is at most $2e^{-d' c/128} \leq 2e^{-d c/128}$.
	
	For the second statement, assume $\OM(y) \leq (1-8\eps)n$, and consider the first vertex $x'$ on the path from $y$ to $x$ such that $\OM(x') \geq (1-6\eps)n$. The probability that more than $\eps n$ bits were flipped in the creation of $x'$ is at most $2^{-\eps n}$ by the Chernoff bound, since by definition of $x'$ the parent of $x'$ has an \OM-value smaller than $(1-6\eps)n$, we may assume that $\OM(x') \leq (1-5\eps)n$. Then, starting from $x'$ we may use the same calculation as above, only that we need to bound the probability that $\eps n$ more zero-bits than one-bits are flipped. This is bounded by the probability that $B_{01} \geq \eps n$. Since $d' \le n/(16 e c)$ we have $\eps n \ge 16 \eps c d' \ge 2 e \E[B_{01}]$, by the Chernoff bound, this probability is at most $2^{-\eps n}$.
	
	\item There can only be one root in a tree, so $s_t^0=1$ for all $t \ge 0$. For $d \ge 1$ and $t \ge 1$, it holds that
	\begin{align*}
		s_t^d = s_{t-1}^d + \sum_{i=1}^{s_{t-1}^{d-1}} Y_i,
	\end{align*}
	where $Y_i$ is an indicator variable that takes value 1 if the $i$-th vertex of depth $d-1$ creates a offspring in round $t$. By Wald's equation, we obtain
	\begin{align*}
		\E[s_t^d] = \E[s_{t-1}^d] + \E[s_{t-1}^{d-1}] / \mu.
	\end{align*}
	Plugging in $\E[s_t^d]=0$ for all $t<d$, we can derive that 
	\begin{equation}
	\label{eq:iteration}
	\E[s_t^d] = \sum_{i=d-1}^{t-1} \E[s_i^{d-1}]/\mu
	\end{equation}
	for all $t \ge d \ge 1$.
	
	We show the result by induction. For $d=1$, by equation (\ref{eq:iteration}) we have $\E[s_t^1] = t/\mu$ for all $t \ge 1$. Now assume that $\E[s_t^d] \le t^d/(d! \mu^d)$ for all $t \ge d$ where $d \ge 1$, again by equation (\ref{eq:iteration}) it holds that
	\begin{align}
	\label{eq:stbound}
		\E[s_t^{d+1}] \le& \sum_{i=d}^{t-1} \frac{i^d}{d! \mu^d} \frac{1}{\mu} = \frac{1}{d! \mu^{d+1}} \sum_{i=0}^{t-1} i^d \nonumber \\
		\le& \frac{1}{d! \mu^{d+1}} \sum_{i=0}^{t-1} t^d 
		= \frac{t^{d+1}}{(d+1)! \mu^{d+1}}
	\end{align}
	for all $t \ge d+1$.
	
	Now consider $t=O(\mu\log\mu)$ and $d=k\log\mu$ for some constant $k > e$. With Stirling's approximation $d!=(1+O(1/d))\sqrt{2\pi d}(d/e)^d$ and equation (\ref{eq:stbound}), we have
	\begin{align*}
		\E[s_t^d] =& O\bigg(\frac{\mu^d (\log \mu)^d}{(d/e)^d \mu^d}\bigg) = O\bigg(\frac{(e \log \mu)^{k \log \mu}}{(k \log \mu)^{k \log \mu}}\bigg) \\ 
		=& O\Big(\frac{\mu^{k}}{\mu^{k \log k}}\Big) = O(\mu^{k(1-\log k)}).
	\end{align*} 
	By Markov's inequality, $\Pr[s_t^d \ge 1] = O(\mu^{k(1-\log k)}\sqrt{\log\mu}) = \mu^{-\Omega(1)}$ as $k(1-\log k) < 0$. 
	Therefore, with probability $1-\mu^{-\Omega(1)}$, $s_t^d=0$ for any $d>e \log\mu$, which implies that the depth of the tree is at most $e \log \mu$.	
	
	For the last statement, let $a_t^d := t^d/(d!\mu^d)$. If $t\ge 2 d \mu$, $a_t^{d-1}/a_t^d = d \mu/t \le 1/2$ for $d \ge 1$. Therefore, 
	\begin{equation*}
		\sum_{i=0}^d \E[s_t^i] \le \sum_{i=0}^d a_t^i \le \sum_{i=0}^d 2^{-(d-i)} a_t^d < 2 a_t^d = \frac{2 t^d}{d! \mu^d}.	\qedhere 
	\end{equation*}
	\end{enumerate}
\end{proof}

\subsubsection{Tail Bound on Steps of $Z_i$}

The first consequence of the coupling is an exponential tail bound on the difference $\best{i}-\best{i-1}$. Note that the tail bound only holds in one direction. There is no comparable tail bound for $\best{i-1}-\best{i}$, at least not without further knowledge on $X_{i-1}$: if there is a single search point $x \in X_{i-1}$ that has $k$ more one-bits than all other search points in $X_{i-1}$, then $x$ might not spawn an offspring and $Z_i$ could drop by $k$ or more, and $k$ could be as large as $\Omega(n)$ without assumptions on $X_{i-1}$.

\begin{lemma}\label{lem:typicalsituation}
	For all $0<\alpha,\eta<1$, $c>0$ there is a constant $\mu_0\in\N$ such that the following holds for all $\mu_0 \leq \mu\leq n$, where $n$ is sufficiently large. Let $i \in [(1-2\eps) \alpha n,(1-\eps/2)\alpha n]$, where $\eps =\mu^{-1+\eta}$. Assume that the \moea with mutation rate $c/n$ on the linear function $f_\ell$ satisfies $\best{i-1} \geq (1-4\eps) n$. Then for all $1\leq \beta \leq \eps n/\log^2\mu$ and $C_2= 6400 e^{\alpha c+1}$,
\begin{align*}
    \Pr[\best{i} - \best{i-1} \geq \beta\cdot C_2\log \mu]\leq 2^{-\beta}.
\end{align*}
If on the other hand $\best{i-1} < (1-4\eps) n$, then $\best{i} < (1-2\eps) n$ with probability $1-e^{-\Omega(\eps n/\log^2\mu)}$.

\end{lemma}

\begin{proof}
By Lemma \ref{lem:findgoodlabel1}, there is $C=16\etac$ such that for all $\beta \ge 1$, 
\begin{align*}
    \Pr[\tdie i - \tbirth i \geq (\beta+2)\cdot C\mu\log \mu]\leq \tfrac14 2^{-\beta}.
\end{align*}

By Lemma \ref{lemma:forest}.1, at round $t = (\beta+2)\cdot C\mu\log \mu$ we have
\begin{align*}
\Pr[s_t \geq \mu^{2(\beta+2) C}]&\leq \frac{t}{\mu^{2(\beta+2) C}} \le (\beta+2) C \mu^{1 - 2(\beta+2)C } \lm < \tfrac14 2^{-\beta},
\end{align*}
where the last step holds for all $\mu \geq \mu_0$ if $\mu_0$ is sufficiently large. That is, the probability that the algorithm visits at least $\mu^{2(\beta+2)C}$ vertices in $X_{\ge i}$ is at most $\tfrac14 2^{-\beta}$.

From now on, we consider $F'$ at a time when it has at most $\mu^{2(\beta+2)C}$ vertices. Let $x$ be a search point that corresponds to a vertex in $F'$ of depth at most $d = \beta C_2' \log \mu$ with root $r$, where $C_2' = 200C/c$. By Lemma \ref{lemma:forest}.2, for $C_2 = 400 e C$ it holds for large enough $\mu_0$ that
\begin{align*}
    \Pr[\text{$x$ and $r$ differ in more than $\beta C_2 \log \mu$ bits}] &\leq 2^{-\beta C_2 \log\mu} \le \tfrac14 2^{-\beta}\cdot \mu^{-2 (\beta+2) C}.
\end{align*}
By a union bound over all vertices in $F'$, the probability that there exists such a vertex $x$ among them is at most $\tfrac14 2^{-\beta}$.

Now let $x$ be a search point that corresponds to a vertex in $F'$ of depth larger than $d$ with root $r$. For large enough $\mu_0$ by Lemma \ref{lemma:forest}.3, if $\OM(r) \geq (1-8\eps)n$ then 
\begin{align*}
    \Pr[\text{$x$ has more one-bits than $r$}]\leq 2 e^{- d c/32} \le \tfrac18 2^{-\beta}\cdot \mu^{-2 (\beta+2) C}.
\end{align*}
The probability that there exists such a vertex $x$ in $F'$ is at most $\tfrac18 2^{-\beta}$ by a union bound. On the other hand, if $n$ is sufficiently large and $\OM(r) \leq (1-8\eps)n$ then for $\beta \le \eps n /\log^{2}\mu$,
\begin{align*}
\Pr[\OM(x) \geq (1-4\eps)n] \leq 2 \cdot 2^{-\eps n} \leq \tfrac18 2^{-\beta}\cdot \mu^{-2 (\beta+2) C},
\end{align*}
Similarly, the probability that such a vertex $x$ exists in $F'$ is at most $\tfrac18 2^{-\beta}$.

To summarize, we have shown that each of the following four events happens with probability at least $1 - 1/4 \cdot 2^{-\beta}$.
\begin{itemize}
	\item $\event{1}$: $T_i - t_i < (\beta + 2) C \mlm$.
	\item $\event{2}$: $s_t < \mu^{2(\beta+2)C} $ at time $t = (\beta + 2) C \mlm$.
	\item $\event{3}$: Among the first $\mu^{2(\beta+2)C}$ vertices in $F'$, there is no search point $x$ with a distance at most $\beta C_2' \lm$ to its root $r$ such that $\left|\{i\in [n] \mid r_i \neq x_i\}\right|>\beta C_2 \log\mu$.
	\item $\event{4}$: Among the first $\mu^{2(\beta+2)C}$ vertices in $F'$, there is no search point $x$ with a distance larger than $\beta C_2' \lm$ to its root $r$ such that either $\OM(r) \geq (1-8\eps)n$ and $\OM(x) > \OM(r)$ or $\OM(r) \leq (1-8\eps)n$ and $\OM(x) \geq (1-4\eps)n$.
\end{itemize} 

Now we argue how the bounds for these events imply the lemma. By $\event{1}$ and $\event{2}$, we may restrict ourselves to the first $\mu^{2(\beta+2)C}$ vertices in $F'$. We claim that there are no offspring $x$ in distance at most $\beta C_2'\log\mu - 1$ from their root $r$ that have \OM-value larger than $\best{i-1} + \beta C_2\log \mu$. To see this, we add the parent of $r$, $r' \in X_{< i}$, and the edge between $r'$ and $r$ to $F'$. Now $r'$ is the root of $x$ and it can act as a reference point: by the definition of $Z_{i-1}$ we have $Z_{i-1} \ge \OM(r')$. If the distance from $r'$ to $x$ is at most $\beta C_2' \lm$, by $\event{3}$ we have $\OM(x) \le \OM(r')+\beta C_2 \lm$. If $x$ is of larger distance from the added root $r'$, we need to discriminate two cases. Either $r'$ has \OM-value at least $(1-8\eps)n$ in which case $\OM(x)$ do not exceed $\OM(r')$ by the first part of $\event{4}$. Or $r'$ has \OM-value at most $(1-8\eps)n$, in which case $x$ do not exceed a \OM-value of $(1-4\eps)n$ by the second part of $ \event{4}$. Therefore, if $\best{i-1} \geq (1-4\eps)n$, we can conclude that \OM-value of $x$ do not exceed $\best{i-1}$ in both cases. Hence, we have shown that $\OM(x) - \best{i-1} > \beta\cdot C_2\log \mu$ is only possible if at least one of the events $\event{1}$ - $\event{4}$ does not occur,  and thus
\begin{align*}
 \Pr[\best{i} - \best{i-1} > \beta\cdot C_2\log \mu]  \leq \sum_{j=1}^4 (1-\Pr[\event{j}]) \leq 2^{-\beta}.
\end{align*}
If $\best{i-1} < (1-4\eps)n$, with the same arguments and letting $\beta \ge \eps n / (\log^2 \mu)$ we have $\best{i} < (1-2\eps) n$ with probability $1-2^{-\Omega(\eps n / \log^2 \mu)}$. \qedhere 

%

\end{proof}

\subsection{Typical Situations}
\label{sec:typical}
As outlined in the overview, our analysis of the drift will be based on studying what happens in 'typical' situations. To characterize these, we use the following definition of 'good' events. Again we consider the \moea on the linear function $f_\ell$. For parameters $\phi,c_d,c_e >0$ we define the event $\Egood(i) := \mathcal E_a \cap \mathcal E_b \cap \ldots \cap \mathcal E_e$, where $\mathcal E_a$ etc. are the following events about the family forest $F_i$ of rank $i$. Recall the family forest consists of all $x \in X_{\geq i}$, and a vertex $u$ is a child of $v$ if $u$ was created as an offspring of $v$. We will be concerned about those vertices in the family forest in $X_i$, i.e., vertices of rank exactly $i$.
\begin{itemize}
    \item $\mathcal E_a$: No vertex in $X_{\le i-1}$ creates offspring in $X_{\ge i+1}$. 
    \item $\mathcal E_b$: There are at most $\eps \mu \log^3\mu$ roots in $F_i$.
    \item $\mathcal E_c$: No vertex in $X_i$ of depth at most $\phi\lm$ in $F_i$ creates offspring in $X_{> i}$. 
    \item $\mathcal E_d$: 
    For every vertex $x\in X_{i}$ that creates an offspring in $X_{\geq i+1}$, if the root $r$ of $x$ has $\OM(r) \geq (1-8\eps)n$ then $\OM(x) \leq \OM(r) - c_d \lm$, and if $\OM(r) \leq (1-8\eps)n$ then $\OM(x) \leq (1-4\eps)n$. Moreover, the mutation changes at most $c_d/2\cdot \lm$ bits.  
    \item $\mathcal E_e$:  No vertex in $X_i$ has an \OM-value which exceeds the \OM-value of its root in $F_i$ by more than $c_e \log \mu$.
\end{itemize}

\begin{lemma}\label{lem:prob_of_good}
For every $0<\alpha<1$, $c>0$ there are $c_d,c_e>0$ such that the following holds. For any constant parameters $0<\phi<1$ and $\eta >0$ that satisfy the following conditions, where $g(\phi) = \phi (\log(8e^{\alpha c+1})- \log\phi)$,
\begin{align}\label{eq:condition_eta}
\eta & <  \min\Big\{g(\phi),\ \frac12-g(\phi),\ \frac{c\phi}{128},\ \frac{c_d}{6}\Big\},
\end{align}
 there exists $\mu_0$ such that for all $\mu_0 \leq \mu\leq n$ and all $i\geq (1-8\eps)\alpha n$, the \moea on $f_\ell$ satisfies
\[
\Pr\left[\Egood(i)\right] \geq 1-O\big(\log^{-2}\mu\big).
\]
\end{lemma}
We remark that $g(\phi) > 0$ for $0 < \phi < 1$ and $g(\phi) < 1/2$ for small enough $\phi$, so there exists $\eta >0$ that satisfies~\eqref{eq:condition_eta}.
\begin{proof}
We need to show that $\Pr[\mathcal E] = 1-O(\log^{-2}\mu)$ holds for $\mathcal E = \mathcal E_a, \ldots, \mathcal E_e$. Thus we split the proof into five parts. Note that we actually show the stronger statement $\Pr[\mathcal E] = 1 - \mu^{-\Omega(1)}$ for $\mathcal E = \mathcal E_a, \mathcal E_c, \mathcal E_d, \mathcal E_e$.

\medskip
\noindent
$\event{a}$:
No vertex in $X_{\le i-1}$ creates offspring in $X_{\ge i+1}$ for $\eta < 1/2$.

We consider the number of offspring that are created from points in $X_{\le i-1}$ and are members of $X_{\ge i+1}$ after the first point $x$ in $X_{\ge i}$ is created. 

We first argue that the probability that $x \in X_i$ is $1-O(\eps)$.  Since we assume $i \ge (1-8\eps) \alpha n$ and the rank of $x$ is at least $i$, the density of zero-bits is $d(A_{\ell+1}, x) \le 8\eps$. By Lemma \ref{lem:Poisson}, 
\begin{equation*}
	\frac{\Pr[x \in X_{\ge i+1}]}{\Pr[x \in X_{\ge i}]} \le 16 \eps \alpha c,
\end{equation*}
which implies $\Pr[x \in X_i \mid x \in X_{\ge i}] = 1 - O(\eps) = 1 - O(\mu^{\eta-1})$.

By Lemma~\ref{lem:findgoodlabel1}, after the first search point in $X_i$ is created, with probability $1-O(\mu^{-\Omega(1)})$ it takes at most $T = 8 e^{\alpha c} \mu\log\mu$ rounds until the set $X_{\leq i}$ is completely deleted. If a search point in $X_{\leq i-1}$ creates an offspring in $X_{\geq i+1}$, at least 2 zero-bits need to be flipped. This probability is $O(\eps^2)$ by a union bound, and hence the expected number of offspring in $X_{\ge i+1}$ created from $X_{\le i-1}$ is at most $O(\eps^2 T) = O(\mu^{-1+2\eta} \log\mu)$. Since $\eta < 1/2$, by Markov's inequality, the probability that the number of such offspring is at least 1 can be bounded by $O(\mu^{-1+2\eta} \log\mu) = O(\log^{-2} \mu)$, as required.

\medskip
\noindent $\event{b}$:
There are at most $\eps \mu \log^3\mu$ roots in $F_i$.

We know from $\event{a}(i-1)$ that we may assume that no points in $X_{i}$ are created from $X_{\leq i-2}$. Hence, it suffices to count the number of roots in $X_i$ that are created from $X_{i-1}$. As in the proof for $\event{a}$, by Lemma~\ref{lem:findgoodlabel1}, after the first search point in $X_{i-1}$ is created, with probability $1-\mu^{-\Omega(1)}$ it takes at most $T = 8 e^{\alpha c}\mu\log\mu$ rounds until the set $X_{\leq {i-1}}$ is completely deleted. In each round we have a probability of at most $p_U = O(\eps)$ to create a new root in $X_i$ ($p_U$ defined in Lemma \ref{lem:Poisson}), so the expected number of roots in $X_i$ is $O(\eps T)$. By Markov's inequality, the number of roots is at most $\eps \mu \log^3\mu$ with probability $O(\eps T/ (\eps \mu \log^3 \mu)) = O(\log^{-2} \mu)$.

\medskip
\noindent
 $\event{c}$: No vertex in $X_i$ of depth at most $\phi\lm$ in $F_i$ creates offspring in $X_{> i}$.

As a sketch for the proof, we first show that the number of vertices of depth at most $\phi\log\mu$ in $F_i$ is at most $\mu^{2g(\phi)}$ with high probability. Then by a simple estimation, the expected number of offspring in $X_{>i}$ created by those vertices is $O(\mu^{{-1+2\eta+2g(\phi)}} \log^4 \mu)$. Since $g(\phi) < 1/2$ for small enough $\phi$, for $\eta < 1/2-g(\phi)$ with probability $1-O(\mu^{-\Omega(1)})$ no such offspring is created.

By Lemma~\ref{lemma:coupling}, we couple $F_i$ with $F'$. Since by $\event{b}$ there are at most $\eps \mu \log^3\mu$ roots in $F_i$, we only need to consider $\eps \mu \log^3\mu$ trees in $F'$. 

Recall that by Lemma~\ref{lem:findgoodlabel1}, the lifetime of $X_i$ is at most $T:=8e^{\alpha c}\mu\log\mu$ with probability at least $1-\mu^{-\Omega(1)}$, if $\mu \geq \mu_0$ for a sufficiently large $\mu_0$. Hence, it suffices to study $F'$ after $T$ rounds. We want to bound the number of vertices with depth at most $\phi\log \mu$. We fix a root, and consider the tree attached to this root. By Property Lemma \ref{lemma:forest}.4 and by the Stirling formula $k! = \Theta(\sqrt{k}(k/e)^k)$ in the second step, the expected number of vertices with depth at most $\phi\log\mu$ at round $T$ is
\begin{align*}
	\sum_{d=0}^{\phi \log\mu} \E[s_T^d] &< 2 \E[s_T^{\phi\log\mu}] = \frac{2 T^{\phi \log\mu}}{(\phi \log\mu)! \mu^{\phi \log \mu}} \\
	&= \Theta\Big(\frac{(8 \etac \mu\log\mu)^{\phi\log\mu}}{\sqrt{\log\mu} (\phi\log\mu/e)^{\phi\log\mu} \mu^{\phi \log\mu}}\Big) \\
	&= o\big(\mu^{\phi(\log(8 e^{\alpha c+1}) -\log\phi)}\big) = o(\mu^{g(\phi)}).
\end{align*}
Note that for $0<\phi < 1$ we have $g(\phi) >0$. By Markov's inequality,
\begin{align*}
	\Pr\bigg[\sum_{d=0}^{\phi\log\mu} s_T^d \ge \mu^{2 g(\phi)}\bigg] = o\big(\mu^{- g(\phi)}\big).
\end{align*}

Since we consider $\eps\mu\log^3\mu = \mu^{\eta}\log^3\mu$ trees in $F'$, by a union bound over all trees, with probability at least $1-o(\mu^{\eta - g (\phi)} \log^3 \mu)$ the number of vertices with depth at most $\phi\log\mu$ is at most $\mu^{\eta+2g(\phi)} \log^3\mu$. Note that the error probability is $o(1)$ since we assumed that $\eta < g(\phi)$.

In each round, every such vertex has a probability of at most $O(\eps/ \mu)$ to create an offspring of strictly larger rank: it must be selected as parent and its offspring must have strictly larger rank. Since the vertices in $X_i$ are present for at most $T=8\etac\mu\log\mu$ rounds, the expected number of offspring in $X_{\ge i+1}$ created by vertices in $X_i$ of depth at most $\phi\log\mu$ is $O(T \cdot \eps /\mu \cdot\mu^{\eta +2g(\phi)} \log^3 \mu)  = O(\mu^{-1 + 2\eta+2g(\phi)} \log^4 \mu)$. By Markov's inequality, the probability that the number of such offspring is at least 1 is $O(\mu^{-1 + 2\eta+2g(\phi)} \log^4 \mu)$. Since $g(x)$ is monotonically increasing in $(0,1)$ and $g(0)=0$, $\eta <  1/2-g(\phi)$ holds for small enough constant $\phi$, making the error probability $\mu^{-\Omega(1)}$. Hence, we have shown that with sufficiently small probability the vertices in depth at most $\phi \lm$ do not create offspring in $X_{> i}$.

\medskip
\noindent
$\event{d}$:
For every vertex $x\in X_{i}$ that creates an offspring in $X_{\geq i+1}$, if the root $r$ of $x$ has $\OM(r) \geq (1-8\eps)n$ then $\OM(x) \leq \OM(r) - c_d \lm$, and if $\OM(r) \leq (1-8\eps)n$ then $\OM(x) \leq (1-4\eps)n$. Moreover, the mutation changes at most $c_d/2\cdot \lm$ bits.

If $\event{c}$ holds, the vertices in $X_i$ that create offspring in $X_{\ge i+1}$ must be of distance at least $d = \phi' \lm$ where $\phi' > \phi$ from their roots. Consider a root $r$ with $\OM(r) \geq (1-8\eps)n$. By equation (\ref{eq:evenmoreonebits}) in Lemma \ref{lemma:forest}, for $c_d=c\phi/16$,  $\Pr[\OM(x)-\OM(r) \geq - c_d\log \mu] \leq 2e^{-c_d \lm/8} = 2\mu^{-M}$, with $M:= c\phi/128$. If $\event{b}(i+1)$ holds, the number of offspring in $X_{\ge i+1}$ created by points in $X_i$ is at most $\eps \mu \log^3 \mu$, which means the number of points in $X_i$ that create offspring in $X_{\ge i+1}$ is at most $\eps \mu \log^3 \mu$. By a union bound, with probability at least $1 - O(\eps\mu \log^3 \mu \cdot 2 \mu^{-M}) = 1 - O(\mu^{-M+\eta} \log^3 \mu)$, a vertex in $X_i$ that creates an offspring in $X_{\ge i+1}$ has a \OM-value which is at least $c_d \lm$ smaller than that of its root. Since we assumed $\eta < M$, this probability is $1-\mu^{-\Omega(1)}$, and thus sufficiently large. This concludes the case that the root has \OM-value at least $(1-8\eps)n$.

If a vertex $x$ has a root which has at most \OM-value $(1-8\eps)n$, we consider the first vertex $x'$ of \OM-value at least $(1-6\eps)n$ on the path from the root to $x$. Then we know that $x'$ has $\OM$-value at most $(1-5\eps)n$, since its direct parent has \OM-value less than $(1-6\eps)n$ and the probability to flip at least $\eps n$ bits in one mutation is $2^{-\eps n}$. Then by similar arguments as above, the probability that a descendant of $x'$ has \OM-value which is $\eps n$ larger than $x'$ is also $2^{-\eps n} = \mu^{-\omega(1)}$, and thus we can easily apply a union bound over $\eps \mu \log^3 \mu$ vertices in $X_i$ that create offspring in $X_{\ge i+1}$.

Finally, we come to the number of bit flips in the improving mutation. In one mutation the expected number of changed bits is $c$. Let $c_d/2 \cdot \lm = (1+\delta')c$ for some $\delta' > 1$, by Chernoff bound, the probability that the number of changed bits is larger than $c_d/2 \cdot \lm$ can be bounded by $e^{-\delta' c/ 3} = \Theta( \mu^{-c_d/6} )$. Similarly, by a union bound, the error probability is at most $O(\eps \mu \log^3 \mu \cdot \mu^{-c_d/6}) =O(\mu^{\eta - c_d / 6} \log^3 \mu)$, which is $\mu^{-\Omega(1)}$ since $\eta < c_d/6$.

\medskip
\noindent
$\event{e}$: No vertex in $X_i$ has an \OM-value which exceeds the \OM-value of its root in $F_i$ by more than $c_e \log \mu$.

We set $c_e := 2 e^2 c k$ where $k>1$ is a positive constant to be chosen later and assume the distance between a vertex $x$ and its root $r$ is $d$. By Lemma \ref{lemma:forest}.4, with probability $1-\mu^{-\Omega(1)}$, $d \le e \log\mu$, and thus $c_e \log\mu \ge 2 e d c$. By Lemma \ref{lemma:forest}.2, the probability that $x$ and $r$ differ in more than $c_e \log\mu$ is at most $2^{-c_e \log\mu} = \mu^{- 2 e^2 c k \log 2}$. Therefore, the probability that $\OM(x)$ exceeds $\OM(y)$ by more than $c_e \log \mu$ is at most $\mu^{- 2 e^2 c k \log 2}$. Moreover, The lifetime of $X_i$ is $8 e^{\alpha c}\mu\log\mu$ with probability $1-\mu^{-\Omega(1)}$ by Lemma \ref{lem:findgoodlabel1}. By Lemma \ref{lemma:forest}.1, with probability $1- O(\mu^{1-e^{\alpha c}} \log \mu)$ there are at most $\mu^{9 e^{\alpha c}}$ vertices in $F_i$. By a union bound over all these vertices, the error probability is at most $O(\mu^{9\etac - 2 e^2 c k \log2})$. By choosing $k > 9\etac / (2 e^2 c \log 2)$, this probability is $\mu^{-\Omega(1)}$. \qedhere
\end{proof}

\subsection{Estimating the Drift}

We are now ready to collect the information to prove negative drift of the $\best i$. We first give a lemma that shows that $\best {i+K}-\best{i}$ is negative in case of good events. As outlined in the introduction, good events don't imply that $\best {i+1}-\best{i}$ is negative, we need to make $K$ steps for some constant $K\in\N$.
\begin{lemma}\label{lem:good_implies_drift}
Let $\ell \in [L]$ and $i\in \N$. Consider the \moea on the linear auxiliary function $f_\ell(x) := n\sum_{j\in \Ae} x_j + \sum_{j\in \Rel} x_j$. Assume that in some step $t \geq 0$ the highest rank in the population is $i$, that $\Egood(i),\ldots,\Egood(i+K)$ hold, where $K := \lceil 2(c_e+1)/c_d\rceil$, and that $\OM(r) \ge (1-8\eps)n$ holds for all roots $r$ in $F_{i},\ldots,F_{i+K-1}$. Then $\best{i+K} \leq \best{i}-\log \mu$.
\end{lemma}
\begin{proof}
Let $j \in \{i+1,\ldots,i+K\}$, and let $r\in X_j$ be any root in $F_j$. By $\mathcal E_a(j-1)$, the parent individual $x$ of $r$ is in $X_{j-1}$. By $\mathcal E_d(j-1)$, the root $r'$ of $x$ in $F_{j-1}$ satisfies $\OM(r') \geq \OM(x) + c_d \log \mu \geq \OM(r) + c_d/2 \cdot \log \mu$. By induction, we obtain that for every root $r\in X_j$ there exists a root $\tilde r \in X_i$ such that $\OM(r) \leq \OM(\tilde r) - (j-i) c_d/2 \cdot \log \mu \leq \best{i} - (j-i) c_d/2 \cdot \log \mu$, where the second step holds since $\OM(\tilde r) \leq \best i$ by definition of $\best i$. Now consider any individual $\tilde x \in X_{i+K}$, and let $r \in X_j$ be its root. By $\mathcal E_e(i+K)$, we have 
\begin{align}\label{eq:good_implies_drift}
\OM(\tilde x) \leq \OM(r) + c_e\log \mu & \leq Z_i - K \cdot c_d/2 \cdot \log \mu + c_e\log \mu \nonumber \\ 
& \leq \best{i}-\log\mu, 
\end{align}
where the latter inequality follows from the definition of $K$. Since \eqref{eq:good_implies_drift} holds for all $\tilde x \in X_{i+K}$, we obtain $Z_{i+K} \leq Z_i-\log \mu$, as required.
\end{proof}

We are now ready to prove the main theorem on the drift of $\best i$. Recall that we have upper, but no lower tail bounds on $\best{i}-\best{i-1}$, cf.\! the comment before Lemma~\ref{lem:typicalsituation}. In order to still be able to apply the negative drift theorem later, we show that the drift is even negative if we truncate the difference $\best{i+K}-\best i$ at $-\log \mu$.

\begin{theorem}\label{thm:truncated_drift}
For every $c>0$ there is a $\mu_0 \in \N$ and a $K \in \N$ such that for all $\mu_0 \le \mu \le n$ where $n$ is sufficiently large the following holds for the \moea with mutation parameter $c$ on the auxiliary function $f_\ell$. Assume that in some generation the fittest search point satisfies~\eqref{eq:conditiondensity}. Then
\begin{align*}
    \E[\max\{\best{i+K} - \best{i},-\log\mu\}] \leq -1.
\end{align*}
\end{theorem}
\begin{proof}
Let $K$ be the constant from Lemma~\ref{lem:good_implies_drift}. Recall from Lemma~\ref{lem:prob_of_good} that the event $\Egood$ has probability $1-O(\log^{-2} \mu)$, which is at least $1/2$ if $\mu$ is sufficiently large. By Lemma~\ref{lem:good_implies_drift}, the event $\Egood$ implies $\best{i+K} - \best{i} \leq - \log \mu$, so in this case the term $\max\{\best{i+K} - \best{i},-\log\mu\}$ evaluates to $-\log \mu$. Hence, let $E_{\text{good}} :=  \E[\max\{\best{i+K} - \best{i},-\log\mu\} \mid \Egood] \cdot \Pr[\Egood]$, it holds that
\begin{align*}
    E_{\text{good}} &= \sum_{j=-\infty}^{\infty}\max\{j,-\log\mu\} \cdot \Pr[\best{i+K}-\best i  = j \land \Egood] \\
    &  = (-\log \mu) \cdot \sum_{j=-\infty}^{\mathclap{-\lceil\log \mu\rceil}}  \Pr[\best{i+K}-\best i = j \land \Egood] \\
    & = -\log \mu \cdot \Pr[\Egood] \leq -2,
\end{align*}
where the second equality holds because $\Pr[Z_{i+K}-Z_{i}=j \land \Egood]=0$ for $j\ge -\floor{\lm}$ and the last step follows from $\Pr[\Egood] \geq 1/2$ if $\mu$ is sufficiently large. 

In the remainder, we will show that the term $E_{\text{good}}$ is very close to $\E[\max\{\best{i+1}-\best i,-\log \mu\}] $. In fact, the difference is
\begin{align}\label{eq:thm_negative_drift}
    &\quad\E[\max\{\best{i+K}-\best i,-\log \mu\}] - E_{\text{good}} \nonumber \\
    &= \sum_{j=-\infty}^{\infty}\max\{j,-\log\mu\} \cdot \Pr[\best{i+K}-\best i = j \land \neg \Egood] \nonumber \\
    &\leq \sum_{j=1}^{\infty}j \cdot \Pr[\best{i+K}-\best i = j \land \neg \Egood].
\end{align}
For an arbitrary constant $C>0$ we may define $j_0 := \lceil C\log\mu\log\log\mu \rceil$. Then we bound $j$ by $j_0$ in the range $j\leq j_0$, and we bound $\Pr[\best{i+K}-\best i = j \land \neg \Egood]$ by $\Pr[\best{i+K}-\best i = j ]$ for $j > j_0$. Since for $j > j_0$,
\begin{align}
\label{eq:upperbound}
	\Pr[Z_{i+K}-Z_i= j] &\le \Pr[Z_{i+K}-Z_i \ge j] \nonumber \\
	&=\Pr\bigg[\sum_{k=1}^K (Z_{i+k}-Z_{i+k-1}) \ge j\bigg] \nonumber \\
	&\le \sum_{k=1}^K \underbrace{\Pr[Z_{i+K}-Z_{i+k-1} \ge j/K]}_{\le 2^{-j/(K C_2 \log\mu)} \text{by Lemma \ref{lem:typicalsituation}}} \nonumber \\
	&\le K 2^{-j / (K C_2 \log \mu)}.
\end{align}
We obtain
\begin{align*}
\eqref{eq:thm_negative_drift} 
&\leq \sum_{j=1}^{j_0}  j_0\cdot \Pr[\best{i+K}-\best i = j \land \neg \Egood]  + \sum_{j=j_0+1}^{\infty}  j\cdot 	\Pr[\best{i+K}-\best i = j] \nonumber\\
&\leq j_0 \cdot \underbrace{\Pr[\neg \Egood]}_{= O(\log^{-2}\mu)} + \sum_{j= j_0+1}^{\infty} j \cdot \underbrace{\Pr[\best{i+1}-\best i = j]}_{\mathclap{\le K 2^{-j / (K C_2 \log \mu)} \text{ by (\ref{eq:upperbound})}}} \\
&= O\big(j_0\log^{-2}\mu\big) +  O\big(\log \mu \cdot j_0 2^{-j_0/(K C_2\log \mu)}\big),
\end{align*}
where the factor $\log \mu$ in the second term appears because $\sum_{s=s_0}^{\infty}s2^{-s/x} = O(xs_02^{-s_0/x})$ for $x\geq 1$, which can be seen by grouping the sum into batches of $x$ summands. The second term is $O(\log^{-1} \mu)$ if we choose the constant $C>0$ in the definition of $j_0 = \lceil C\log\mu\log\log\mu \rceil$ appropriately. The first term is $O(\log \log \mu \cdot \log^{-1} \mu)$. Hence, by choosing $\mu$ sufficiently large, we can make both terms smaller than $1/2$, and obtain that $\E[\max\{\best{i+K}-\best i,-\log\mu\}] \leq E_{\text{good}} +1 \leq - 1$, as desired. \qedhere
\end{proof}

\section{Proof of Theorem~\ref{thm:main}}
\label{sec:proofthm}

In the previous section we have analyzed the random variable $\best i$, and in particular we have shown that it has negative drift. In this section we will show how our main result, the lower bound on the runtime for the \moea, follows from the negative drift of $\best i$. The proof follows from similar ideas as in~\cite{lengler2018drift} and~\cite{lengler2018general}. We start with a lemma that describes the behavior of the \moea on $f_{\ell}$.

\begin{lemma}\label{lem:fixed_level}
For every constant $0<\delta <2/7$ the following holds. Let $\ell \in [L]$ and consider the \moea on $f_{\ell}$ under the assumption that $d([n],x) \geq \eps(1+2\delta)$ and $d(A_{\ell+1},x) \ge \eps(1+\delta) $  
hold for all $x$ in the initial population. For $t\geq 0$, let $x^{t}$ be the offspring in round $t$.  Then with probability $1-\exp\{-\Omega(\eps n/\log^2 \mu)\}$, the following holds for all $t \leq L$.
\begin{itemize}
\item $d([n],x^{t}) \geq \eps(1+\delta)$.
\item $d([n],x^{t}) \geq \eps(1+2\delta)$ or $d( A_{\ell+1},x^{t}) \geq \eps(1+\delta/4)$.
\end{itemize} 
\end{lemma}
Before we come to the proof, let us briefly explain why the lemma is useful. It is tailored to support an inductive proof for Theorem~\ref{thm:main} for \hottopic. In this induction, we will show that $d([n],x^{t}) \geq \eps(1+\delta)$ for exponential time. In fact, when the algorithm enters a new level then the density is at least $\eps(1+2\delta)$. Moreover, one can show that with high probability the new hot topic did not influence the algorithm up to this point, so it behaves just as a random subset of positions of size $\alpha n$. In particular, with high probability its density is at least $\eps(1+\delta)$, so the assumptions of the lemma are satisfied. As long as the level does not change, the \hottopic function is identical to $f_\ell$, so we may apply Lemma~\ref{lem:fixed_level}. The first item implies what we actually want to prove, at least as long as we stay on the same level. For the second item, by the construction of the \hottopic function the level increases when $d( A_{\ell+1},x^{t}) \approx \eps < \eps(1+\delta/4)$. So the second item implies that at this point in time we have $d([n],x^{t}) \geq \eps(1+2\delta)$, which is the requirement for the next step in the induction. Note that we can't just merge the items into one. For example, if we would weaken the second item to assert $d([n],x^{t}) \geq \eps(1+\delta)$ at the beginning of a level, then we could not conclude that the next offspring satisfies the same bound with exponentially small error probability.
\begin{proof}[Proof of Lemma~\ref{lem:fixed_level}]
Let $i_0$ be the largest rank in the initial population, i.e., the largest number of one-bits in $A_{\ell+1}$ in the initial population. We fix an offset $a\in\{0,\ldots,K-1\}$ and consider the sequence of random variables $Y_{i,a} := \best{i_0+a+iK}/\log \mu$, where $i$ is a non-negative integer. In the initial population, each individual has at most $n\left(1-\eps(1+2\delta)\right)$ one-bits by assumption. Hence, we also have $\best{i_0+a} \le n(1-\eps(1 + 3\delta/2))$ with probability $1-\exp\{-\Omega(\eps n)\}$ for all offsets $a\in\{0,\ldots,K-1\}$, since otherwise at least one of the $K$ mutations would need to flip $\Omega(\eps n)$ bits, which happens only with probability $\exp\{-\Omega(\eps n)\}$ by the Chernoff bound. Thus for the first statement it suffices to show that $Y_{i,a} \le Y_{0,a} + \eps\delta n/(2\log\mu)$ for all $i \geq 0$. Since that is equivalent to $\best{i_0+a+iK} \le \best{i_0+a} + \eps\delta n/2$ for all $i \ge 0$, and we already have $\best{i_0+a} \le n(1-\eps(1 + 3\delta/2))$ for all $a$ with high probability, altogether it implies $\best{i'} \le n(1-\eps(1 + \delta))$ for all $i' \ge i_0$. As $Z_{i'}$ denotes the maximum number of one-bits in rank $i'$, we conclude that $d([n],x^{t}) \geq \eps(1+\delta)$ holds for any individual $x^t$ of rank $i' \ge i_0$. For the second statement, we distinguish between two cases. Note that the index $i$ counts, up to the factor $K$, the increase in one-bits in $A_{\ell+1}$. If $i \le \alpha n \eps\delta/(4K) - 1$, then for any $x^t$ of rank $i_0+a+i K$, $d( A_{\ell+1},x^{t}) \ge (\alpha n \eps(1+\delta/2) - a - i K)/(\alpha n) > \eps(1+\delta/2) - (i+1)K/(\alpha n) \ge \eps(1+\delta/4)$. For $i > \alpha n \eps\delta/(4K) - 1$, we aim to show that $Y_{i,a} \le Y_{0, a}$.

We would like to apply the negative drift theorem to $Y_{i,a}$ for the range $[(1-\eps(1+3\delta/2))n/\log\mu, (1-\eps(1+\delta))n/\log\mu]$. First note that we study a linear function, and that the bits in $A_\ell$ have larger weights than the remaining bits. Thus, it can be shown by a coupling argument (Lemma 4.2 in~\cite{lengler2018drift}) that if $d(A_{\ell+1},x) \le d([n],x)+\delta\eps$ holds initially, then the slightly weaker condition $d(A_{\ell+1},x) \le d([n],x)+2\delta\eps $ remains true for all individuals in the population for the next $L$ rounds, with probability at least $1-L e^{-\Omega(\eps n)}$. By choosing the constant parameter $\rho$ in the definition of $L = \exp\{\rho \eps n/\log^2 \mu\}$ small enough, the factor $L$ can be swallowed by the term $e^{-\Omega(\eps n)}$. Thus we may assume that whenever $Y_{i,a}$ is in the range $[(1-\eps(1+3\delta/2))n/\log\mu, (1-\eps(1+\delta))n/\log\mu]$ then $d(A_{\ell+1},x) \leq d([n],x)+2\delta\eps \leq \eps (1 + 3\delta/2+2\delta) \leq 2\eps$ as $\delta < 2/7$. In addition, we have $d(A_{\ell+1}, x) \ge \eps/2$ before the level changes, since otherwise with probability $1-e^{-\Omega(\eps n)}$ it holds that $d(B_{\ell+1}, x) < \eps$, which implies an increase of level. Thus the conditions in (\ref{eq:conditiondensity}) are satisfied, and thus Lemma 6 is applicable.

So let us study the drift of $Y_{i,a}$ in the range $[(1-\eps(1+3\delta/2))n/\log\mu, \\ (1-\eps(1+\delta))n/\log\mu]$. First note that the probability to jump over more than half of this interval is $\exp\{-\Omega(\eps n/\log^2\mu)\}$: for $Y_{i,a} \geq (1-4\eps)n/\lm$ this follows from the first statement in Lemma~\ref{lem:typicalsituation}, for $Y_{i,a}< (1-4\eps)n/\lm$ it follows from the second statement in Lemma~\ref{lem:typicalsituation}. So we may assume that $Y_{i,a}$ is contained in the first half of the interval for some $i=i^*$. To ease notation, we will assume $i^*=0$. Inside of the interval, by Theorem~\ref{thm:truncated_drift}, it holds that
\begin{equation*}
\E[Y_{i+1,a}-Y_{i,a}] = \E[Z_{i_0+a+(i+1)k}-Z_{i_0+a+i k}]/{\log \mu} \le -1/{\log\mu}.
\end{equation*}
Moreover, by Lemma~\ref{lem:typicalsituation} the sequence of random variables $(Y_{i,a})_{i\geq 0}$ has an upper exponential tail bound, i.e., $\Pr[Y_{i+1,a} -Y_{i,a} \ge K \cdot \beta C_2 ] \leq K \cdot 2^{-\beta}$ for all $1 \le \beta \le \eps n / \log^2 \mu$. (In particular, the probability that there is ever a jump larger than $K C_2 \eps n/\log^2\mu$ within $L$ steps is at most $O(L \cdot 2^{-\eps n/\log^2\mu}) = o(1)$, so we may assume that such jumps never occur.) To show sub-Gaussianity, we should extend the inequality also for $\beta<1$. Since any probability is bounded by 1, the bound $\Pr[Y_{i+1,a}-Y_{i,a}+1/{\log\mu} \ge K \beta C_2 + 1/{\log\mu}] \le 2 K\cdot 2^{-\beta}$ is not just true for $\beta \geq 1$, but also trivially satisfied for any $\beta \in [0,1]$. Therefore, for any $y\ge 0$, it holds that
\begin{align*}
\Pr[Y_{i+1,a}-Y_{i,a}+1/{\log\mu} \ge y] & \le 2^{1+1/(K C_2 \log\mu)} K (2^{1/(K C_2)})^{-y} \nonumber \\
&\le 2^{1+1/(K C_2 \log2)}K (2^{1/(K C_2)})^{-y}.
\end{align*}

However, we need exponential tail bounds in both directions, so we need to truncate the downwards steps of $Y_{i,a}$ as follows. We set $\tilde Y_{0,a} := Y_{0,a}$, and we define $\tilde Y_{i,a}$ recursively by $\tilde Y_{i,a} -\tilde Y_{i+1,a} := \min\{Y_{i,a} -Y_{i+1,a}, 1/{\log\mu}\}$. Then clearly we have $\tilde Y_{i,a} \geq Y_{i,a}$ for all $i \geq 0$, and $\tilde Y_{i,a}$ satisfies the tail bound condition that 
\begin{equation*}
\Pr[\tilde Y_{i,a}-\tilde Y_{i+1,a}-1/{\log\mu} > 0] = 0.
\end{equation*}
Therefore, by Theorem~\ref{thm:subgaussian}, $(\tilde Y_{i,a} + i/{\log\mu})_{i \ge 0}$ is $(128 c' \delta'^{-3}, \delta'/4)$-sub-Gaussian, where $c'=2^{1+1/(K C_2 \log 2)} K$ and $\delta'=2^{1/(K C_2)}-1$. And by Theorem~\ref{thm:bound},
\begin{equation*}
\Pr\Big[\max_{0\le j\le i}(Y_{j,a}-Y_{0,a}) \ge -i/{\log\mu}+y\Big] \le \exp\Big(-\frac{\delta' y}{8}\min\Big(1, \frac{\delta'^2}{32c'} \cdot \frac{y}{i}\Big)\Big).
\end{equation*}

Now for any $i \ge 0$ and $y=i/{\log\mu}+\eps\delta n/(4\log\mu)$, with probability $1-\exp\{-\Omega(\eps n/\log^2 \mu)\}$ we have $Y_{i,a} \le Y_{0,a}+\eps\delta n/(4\log\mu)$. Note that $\eps\delta n/(4\log\mu)$ is half of the length of the interval of interest, which implies that $Y_{i,a}$ does not go beyond the interval with high probability. Similarly, for every fixed $i \ge \alpha n \eps\delta/(4K)$ and $y = i/{\log\mu}$ we have $Y_{i,a} \le Y_{0,a}$ with probability $1-\exp\{-\Omega(\eps n/\log^2 \mu)\}$. The proof is concluded by a union bound over all possible $i$. Since there are at most $n$ possible values, this increases the error probability by a factor of $n$, which we can swallow in the expression $\exp\{-\Omega(\eps n/\log^2 \mu)\}$.
\end{proof}

Finally, we have collected all ingredients to prove our main result.

\begin{proof}[Proof of Theorem~\ref{thm:main}]
 Let $L := \exp\{\rho \eps n/\log^2\mu\}$ be the number of levels. For the proof, we will consider an auxiliary run of the \moea with a dynamic fitness function $\tilde f$ in which we only allow the levels to increase by one. In particular, the function $\tilde f$ does not only depend on the current state of the algorithm, but also on the algorithm's history. More precisely, we define an auxiliary level $\tilde \ell(x,t)$ of a search point $x$, which we only allow to increase by at most one per round. Recall that~$\ell(x)$ was defined in~\eqref{eq:level} as $\ell(x) = \max \{ \ell' \in [L] : d(B_{\ell'},x) \le \eps\}$. For $\tilde \ell(t)$, we use the same definition except that we let the maximum go over only $\ell'  \leq \min\{\tilde \ell(t-1)+1, L\}$. I.e., we set $\tilde \ell(0) := 0$, and if an offspring $y^t$ of $x^{t}$ enters the population in round $t$, then we set $\tilde \ell(y^t, t) := \max \{ \ell' \in [\min\{\tilde \ell(x^t, t-1)+1,L\}] : d(B_{\ell'},y^{t}) \le \eps\}$. (If the population stays the same in round $t$, then we leave $\tilde \ell$ unchanged.) Then we define the auxiliary fitness of $y^t$ as
  \begin{align*}
\tilde f(y^t) :=\tilde \ell(y^t, t) \cdot n^{2} \ +  \sum_{\mathclap{{i\in A_{\tilde \ell(y^t, t)+1}}}}  y^t_i\cdot n \ +   \sum_{\mathclap{{i\in R_{\tilde \ell(y^t, t)+1}}}}  y^t_i , 
\end{align*}
i.e., we use the same definition as for the \hottopic function except that we replace $\ell(y^t)$ by $\tilde \ell(t)$. Then we proceed as the \moea, i.e., in each round we compute and store the auxiliary fitness of the new offspring (which may depend on the whole history of the algorithm), and we remove the search point for which we have stored the lowest auxiliary fitness. This definition does not make much sense from an algorithmic perspective, but we will see in hindsight that the auxiliary process behaves identical to the actual \moea. We will next argue why this is the case.

For the auxiliary process, it is obvious that we only need to uncover the set $A_{i+1}$ and $B_{i+1}$ when we reach level $\tilde \ell(t) =i$. As we will show later for the auxiliary process, with high probability the density $d([n],x^{t})$ stays strictly above $\eps\cdot (1+\delta)$ for a suitable constant $\delta >0$. Now fix any round $t$ with auxiliary level $\tilde \ell(t)$. Since we do need to uncover $B_{\tilde \ell(t) +2}$ at some point after time $t$, its choice does not influence the behavior of the auxiliary process until time $t$. Hence, we can first let the auxiliary process run until time $t$, and afterwards uncover the set $B_{\tilde \ell(t) +2}$. Since $B_{\tilde \ell(t) +2} \subset [n]$ is a uniformly random subset of size $\beta n$, it contains at least $\beta \eps (1+\delta) n$ zero-bits in expectation, and the probability that $B_{\tilde \ell(t) +2}$ contains at most $\beta \eps n$ zero-bits is $\exp\{-\Omega(\beta \eps n)\}$. The same argument also holds for $B_{\tilde \ell(t)+3}, \ldots, B_L$. Since $L = \exp\{\rho \eps n/\log^2 \mu\}$ with desirably small $\rho>0$, we can afford a union bound over all such sets and all times $t \leq L$, which is a union bound over less than $L^2 = \exp\{2\rho \eps n/\log^2 \mu\}$ terms. Hence, with high probability we have $d(B_i, x^{t})> \eps$ for all $1\leq t\leq L$ and all $\tilde \ell(t) +2 \leq i \leq L$. A straightforward induction shows that this implies $\ell(t) = \tilde \ell(t)$ for all $t \leq L$, and thus the \moea behaves identical to the auxiliary process. Note that this already implies that the \moea visits each of the $L$ levels, which implies the desired runtime bound. It only remains to show that there is a constant $\delta >0$ such that the auxiliary process satisfies $d([n],x^{t}) > \eps\cdot (1+\delta)$ for all $t \le L$.

The advantage of the auxiliary process is that we may postpone drawing $A_{\ell+1}$ until we reach level $\tilde \ell = \ell$. In particular, since $A_{\ell+1} \subseteq [n]$ is a uniformly random subset, we may use the same argument as before and conclude that $|d(A_{\ell+1},x) - d([n],x)| < \delta \eps$ holds with probability $1-\exp\{-\Omega(\eps n)\}$ for any constant $\delta >0$ that we desire, and for all members $x$ of the population when we reach level $\ell$. In fact, we have exponentially small error probability, so we may afford a union bound and conclude that with high probability the same holds for all $\ell$. We want to show that the auxiliary process, if running on level $\ell$ and starting with a population that initially satisfies $|d(A_{\ell+1},x) - d([n],x)| < \delta \eps$ for $\delta<2/7$, maintains $d([n],x^{t}) \geq \eps(1+\delta)$ for all new search points $x^{t}$ until $t > L$. 

By the first conclusion from Lemma~\ref{lem:fixed_level}, $d([n],x^{t}) \geq \eps(1+\delta)$ holds as long as the level remains to be $\ell$ and $t \le L$. When a point $x$ reaches level $\ell+1$, by definition we have $d(B_{\ell+1}, x) < \eps$. Since $B_{\ell+1}$ is a uniformly random subset of $A_{\ell+1}$, by the Chernoff bound $d(A_{\ell+1}, x) < \eps(1+\delta/4)$ holds with probability $1-\exp\{-\Omega(\eps n)\}$. So we apply the second conclusion of Lemma~\ref{lem:fixed_level} to $x$ and conclude that $d([n], x) \ge \eps(1+2\delta)$. With high probability, it holds that $d(A_{\ell+2}, x) \ge \eps(1+2\delta) - \eps\delta$ and the conditions in Lemma~\ref{lem:fixed_level} are satisfied again for level $\ell+1$. By induction we obtain $d([n],x^{t}) \geq \eps(1+\delta)$ for all $t \le L$. As the choice of $\ell$ is arbitrary, we start with $\ell=0$ and $d([n],x^t) \ge \eps(1+\delta)$ holds for all $t \le L$. This concludes the proof. \qedhere

\end{proof}

\section{Simulation}
\label{sec:simulation}
In this section we will illustrate the detrimental effect of large populations on $(\mu+1)$-EAs by numerical simulations. Unless otherwise stated, the parameters that we used to generate the \hottopic functions are $n=10000$, $L=100$, $\alpha=0.25$, $\beta=0.05$ and $\eps=0.05$. Each data point is obtained by 10 independent runs on the same \hottopic function with different random seeds. Our implementation is available at \url{https://github.com/zuxu/MuOneEA-HotTopic}.

\subsection{Population Size}
\begin{figure}[h]
\centering
\includegraphics[width=\textwidth]{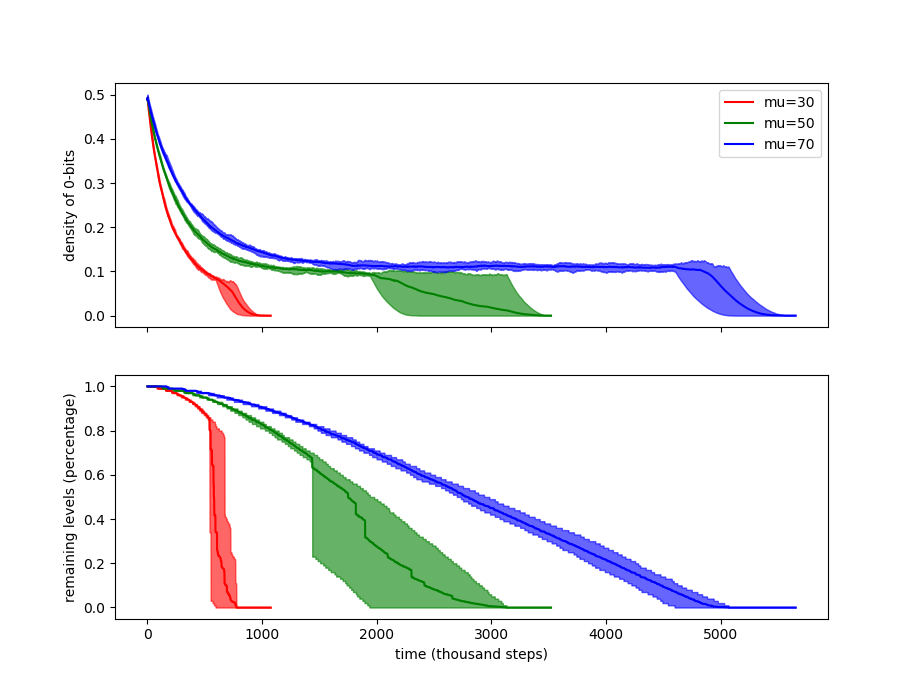}
\caption{Distance to the optimum as $(\mu+1)$-EAs with $c=1.0$ proceed. The solid lines are the mean of 10 simulations, while the shaded areas are bounded by corresponding minimum and maxmium values at each time step.}
\label{fig:curve}
\end{figure}
First of all, we plot typical behaviours of evolutionary algorithms with small, medium and large population sizes. Figure \ref{fig:curve} shows the distance between the optimum and the fittest point $p^*$ in the population with respect to time. We have two metrics for the distance: the density of 0-bits in $p^*$ and the remaining levels of $p^*$ divided by $L$. As indicated by the sudden drops in level, for a small population size ($\mu = 30$), the algorithm skips many levels and reaches the optimum quickly. In contrast, an algorithm with large $\mu = 70$ visits the levels one by one, without improvement on the fitness. This happens where the density of 1-bits is relatively high, such that even though it gradually improves on the current hot topic, it often accepts offspring that flip 1-bits to 0-bits outside of the hot topic. With such offspring accumulating in a large population, the average density of 0-bits remains significantly above $\eps$ before reaching the last level. Therefore, with high probability the algorithm does not skip any level. Once the highest level is reached, the remaining bits can be optimized easily as in the coupon collector. For $\mu=50$, the density gets close, but slightly above $\eps$, so that it depends on chance whether levels are skipped or not. This leads to a high variance in the running time.

\begin{figure}[h]
\centering
\includegraphics[width=\textwidth]{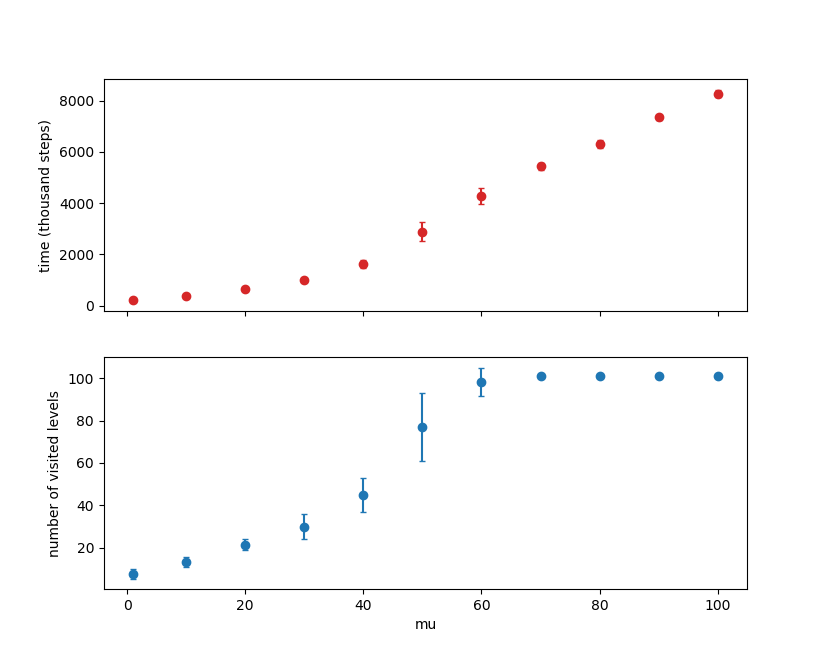}
\caption{Running time and number of visited levels for $(\mu+1)$-EAs with different values of $\mu$ and $c=1.0$. Solid dots indicate the means and error bars show the standard deviations.}
\label{fig:mu}
\end{figure}

In Figure \ref{fig:mu}, we show the running time and the number of visited levels for a wide range of $\mu$. The running time is highly concentrated when $\mu$ is very small or very large. The reason is that the algorithm keeps skipping levels with small $\mu$ and visits all levels with large $\mu$. For a medium sized $\mu$ like 50, level skipping only happens a few times. Since each time when the algorithm skips a level, it lands at some higher level uniformly at random due to the definition of the \hottopic function, which results in a larger variance in the running time.        

\subsection{Mutation Rate}
\begin{figure}[h]
	\centering
	\includegraphics[width=\textwidth]{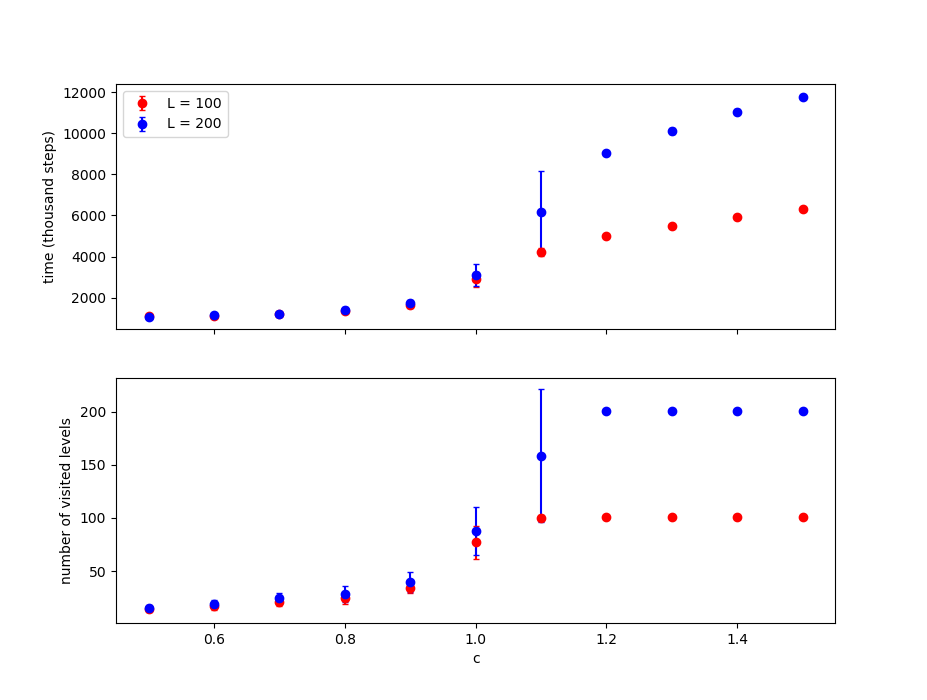}
	\caption{Running time and number of visited levels for $(\mu+1)$-EAs with different values of $c$ and $\mu=50$. Solid dots indicate the means and error bars show the standard deviations.}
	\label{fig:c}
\end{figure}
The mutation rate $c$ is the other factor that affects the magnitude of the negative drift, so we also plot the running time for various values of $c$, see Figure \ref{fig:c}. For a small $c$, detrimental mutations do not occur frequently and thus the average density of 1-bits in the population keeps increasing. Conversely, with a large $c$, the algorithm tends to visit all the levels. To demonstrate the resulting effect on the running time, we compare the cases where $L=100$ and $L=200$. If $c$ is small ($c\leq 0.9$), the algorithm skips levels quickly, and the running time is almost independent of the number of levels. On the other hand, if $c$ is large ($c\geq 1.2$) then the algorithm visits every level. In this case, the running time is essentially proportional to the number of levels, plus some initial phase. Note that in this range the running time can get almost arbitrarily bad, since doubling the number of levels $L$ will essentially lead to a doubling of the running time. As our theoretical analysis shows, this holds even when $L$ becomes exponential in $n$, but for so many levels the running time becomes too large to run experiments. Finally, for a medium sized $c$ like 1.1, level skipping only happens a few times. Each time when the algorithm skips a level, it lands at some higher level uniformly at random, which results in a larger variance in the running time, similar to Figure~\ref{fig:mu}.

\begin{figure}[h]
	\centering
	\includegraphics[width=\textwidth]{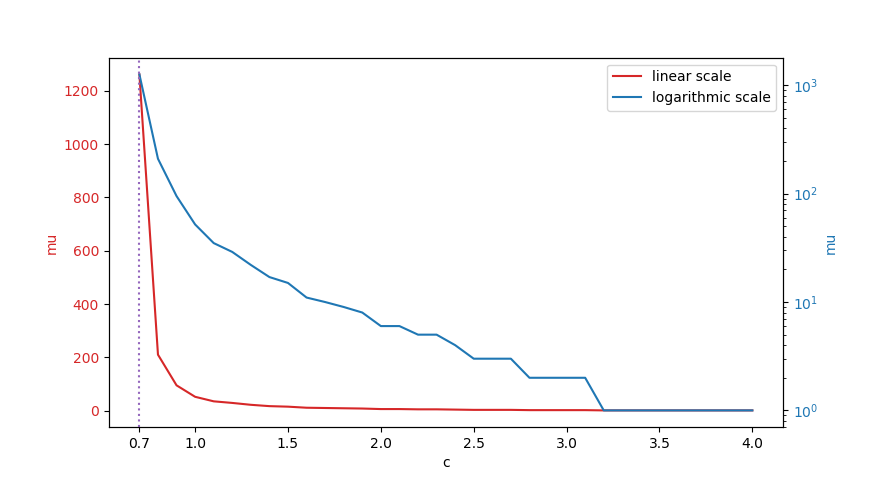}
	\caption{Minimum values of $\mu$ with respect to $c$ such that the \moea visits all levels in at least 5 out of 10 simulations. The choices of $c$ ranges from $0.7$ to $4.0$ with step size $0.1$.}
	\label{fig:muc}
\end{figure}
Finally, we investigate how values of $\mu$ and $c$ jointly influence the behaviour of a \moea. For a fixed $c$, we search for the minimum value of $\mu$ such that the algorithm visits all levels in at least half of the 10 simulations. That is, we seek for a minimum $\mu$ that induces long running times with constant probability. As we can see from Figure \ref{fig:muc}, large values of $c$ ($\ge 3.2$) are extremely harmful even when there is only one individual in the population. An algorithm with a large population can benefit greatly from having a small mutation rate. We did not observe a stable slowdown for $c=0.7$ until we raise the value of $\mu$ to more than $1000$.

\section{Conclusion}
\label{sec:conclusion}

We have shown that the \moea with arbitrary mutation parameter $c>0$ needs exponential time on some monotone functions if $\mu$ is too large. This is one of the very few known situations in which even a slightly larger population size $\mu$ can lead to a drastic decrease in performance. The main reason is that, if progress is steady enough that the population does not degenerate, the search points that produce offspring are typically not the fittest ones. We believe that this is an interesting phenomenon which deserves further investigations, also in less artificial contexts. 

For example, consider the \moea on weighted linear functions with a skewed distribution (e.g., on \Binval), and with a fixed time budget (so that the action happens away from the optimum). It is quite conceivable that the same effect hurts performance, i.e., if the algorithm flips a high-weight bit, it will allow (almost) any offspring of this individual into the population, even though this offspring has probably fewer correct bits than other search points in the population. Does that mean that the fixed-budget performance of the \moea on \Binval deteriorates with increasing $\mu$? Are the resulting individuals further away from the optimum?

An even more pressing question is about crossover. We have studied the \moea, but do the same results also apply for the \moga? In~\cite{lengler2018general} it was shown that close to the optimum (for small values of the \hottopic parameter $\eps$) crossover helps dramatically, and that a large population size can even counterbalance large mutation parameters $c$. So, close to the optimum, for the \moga the effect of large population size was beneficial, while for the \moea it was neutral and did not affect the threshold $c_0$. Thus if we study the \moga on \hottopic functions with large $\eps$, then a beneficial effect of large populations is competing with a detrimental effect. Understanding this interplay would be a major step towards a better understanding of crossover in general.

Similarly, since the problems originate in non-trivial populations, what happens if we equip the \moea with a diversity mechanism (duplication avoidance, genotypical or phenotypical niching), and study it close to the optimum? Does it fall for the same traps? This question was already asked in~\cite{lengler2018general}, but our results shed additional light on the question.

Finally, it is open whether the \moea is fast on any monotone function if it starts close enough to the optimum. i.e., for every $\mu\in \N$, does there exist an $\eps= \eps(\mu)$ such that the \moea, initialized with a random search point with $\eps n$ zero-bits, has runtime $O(n \log n)$ for every monotone function? Of course, the same question also applies to other algorithms like the \moga and the `fast' counterparts of the \moea and the \moga. Interestingly, the result in~\cite{lengler2018general} that the `fast \olea' with good parameters is efficient for every monotone function was only proven under this assumption, that the algorithm starts close to the optimum. So this also raises the question whether there are traps for the `fast \olea' that only take effect far away from the optimum.


\end{document}